\documentclass[letterpaper, 10 pt, conference]{ieeeconf}  % Comment this line out if you need a4paper

\IEEEoverridecommandlockouts                              % This command is only needed if 
\usepackage{amsmath,amssymb,amsfonts}
\usepackage{algorithmic}
\usepackage{graphicx}
\usepackage{subcaption}  
\usepackage{textcomp}
\usepackage{xcolor}
\usepackage{multirow}
\usepackage{makecell}
\usepackage{balance}
\usepackage[font=small]{caption} 
\def\BibTeX{{\rm B\kern-.05em{\sc i\kern-.025em b}\kern-.08em
    T\kern-.1667em\lower.7ex\hbox{E}\kern-.125emX}}
    
\begin{document}

\title{\LARGE \bf ContraMap: Contrastive Uncertainty Mapping for\\ Robot Environment Representation
}
\author{Chi Cuong Le \and Weiming Zhi \thanks{W. Zhi is with the School of Computer Science and the Australian Centre for Robotics, University of Sydney, Australia.}}
\maketitle

\begin{abstract}
Reliable robot perception requires not only predicting scene structure, but also identifying where predictions should be treated as unreliable due to sparse or missing observations. We present \textbf{ContraMap}, a contrastive continuous mapping method that augments kernel-based discriminative maps with an explicit uncertainty class trained using synthetic noise samples. This formulation treats unobserved regions as a contrastive class, enabling joint environment prediction and spatial uncertainty estimation in real time without Bayesian inference. Under a simple mixture-model view, we show that the probability assigned to the uncertainty class is a monotonic function of a distance-aware uncertainty surrogate. Experiments in 2D occupancy mapping, 3D semantic mapping, and tabletop scene reconstruction show that ContraMap preserves mapping quality, produces spatially coherent uncertainty estimates, and is substantially more efficient than Bayesian kernel-map baselines.
\end{abstract}

\section{Introduction}

Reliable robot operation depends on scene representations that are both continuous and uncertainty-aware. Classical grid-based methods, such as occupancy grids \cite{OccupancyGridMaps}, discretise space into fixed cells, which limits spatial resolution and scalability. Modern continuous formulations instead model occupancy or semantics directly over continuous space using Gaussian Processes or kernel-feature methods \cite{GPOM,HilbertMaps}. These representations support high-fidelity mapping and efficient spatial querying, but they typically either omit uncertainty or rely on Bayesian inference, whose computational cost can become prohibitive for large-scale or real-time deployment.

The core challenge is not only to predict occupancy or semantics continuously, but also to represent \emph{where the map should abstain from confidence} in sparsely observed or unobserved space, without incurring the cost of posterior inference. This is particularly important in robotics, where overconfident predictions in occluded or unexplored regions can degrade downstream planning and decision-making. The quality of the map along with estimates of it's uncertainty can be leveraged for further downstream motion generation \cite{Diff_templates,GeoFab_gloabL_opt, Fast_diff_int}. Here, We introduce \textbf{ContraMap}, \emph{Contrastive Uncertainty Mapping}, a continuous kernel-based mapping method that models unobserved space as an explicit uncertainty class. ContraMap augments a standard softmax mapping model with synthetic contrastive samples labelled as ``uncertain'', enabling the model to jointly predict environment structure and a spatial uncertainty score. In occupancy mapping, this captures the transition between free, occupied, and unobserved space; in semantic mapping, it highlights ambiguous or weakly supported regions around objects. The method retains the linear-time optimisation and inference efficiency of discriminative kernel maps while avoiding Bayesian posterior inference.

The main contributions of this work are:
\begin{enumerate}
\item {A contrastive formulation for continuous uncertainty-aware mapping} that represents unobserved regions as an explicit uncertainty class within a kernel-based discriminative map.
\item {A theoretical characterisation} showing that, under a simple mixture-model assumption, the probability assigned to the uncertainty class is a monotonic function of a distance-aware uncertainty surrogate.
\item {Experimental validation} across 2D occupancy mapping, 3D semantic mapping, and tabletop scene reconstruction showing competitive mapping quality and substantial efficiency gains over Bayesian kernel-map baselines.
\end{enumerate}
% \begin{figure}[t]
%     \centering
%     \includegraphics[width=1.0\columnwidth]{figures/teaser.pdf}
%     \caption{Our method employs the softmax architecture with an additional "uncertain" output class. Trained on augmented data that includes noise, our model is able to represent the scene accurately, while simultaneously estimating uncertainty.}
%     \label{fig:teaser}
% \end{figure}

The paper is organized as follows. Section \ref{sec:related_works} discusses related previous studies. In Section \ref{sec:2}, we review the Bayesian methods used in representing different kind of environments, and review about noise contrastive estimation, which is an important cornerstone of our approach. Next, Section \ref{sec:3} present an overview of the proposed method. After that, we present theoretical and empirical insight into our proposed approach. Our experiments demonstrate the effectiveness of proposed strategy are shown in Section \ref{sec:5}. Finally, Section \ref{sec:7} concludes our paper.

\begin{figure}
    \centering
    \includegraphics[width=1.0\linewidth]{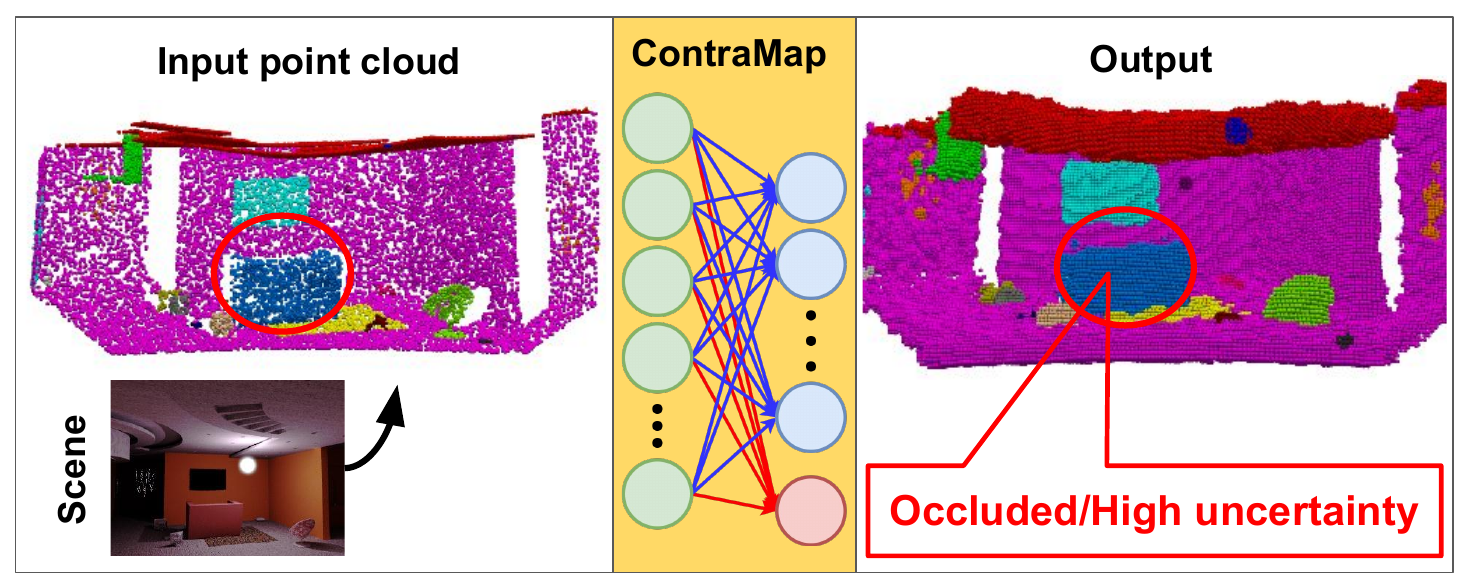}
    \caption{For robust robot operation, scene representations should provide not only spatially consistent mapping but also a measure of uncertainty across the environment. \textbf{ContraMap} augments continuous classification-based mapping with an additional uncertainty output, enabling joint environment representation and direct uncertainty prediction at any queried location. The model reconstructs scene structure while assigning high uncertainty to occluded or weakly observed regions, such as the space behind the table.}
    \label{fig:teaser}
\end{figure}

\begin{figure*}[t]
    \centering
    \includegraphics[width=0.85\linewidth]{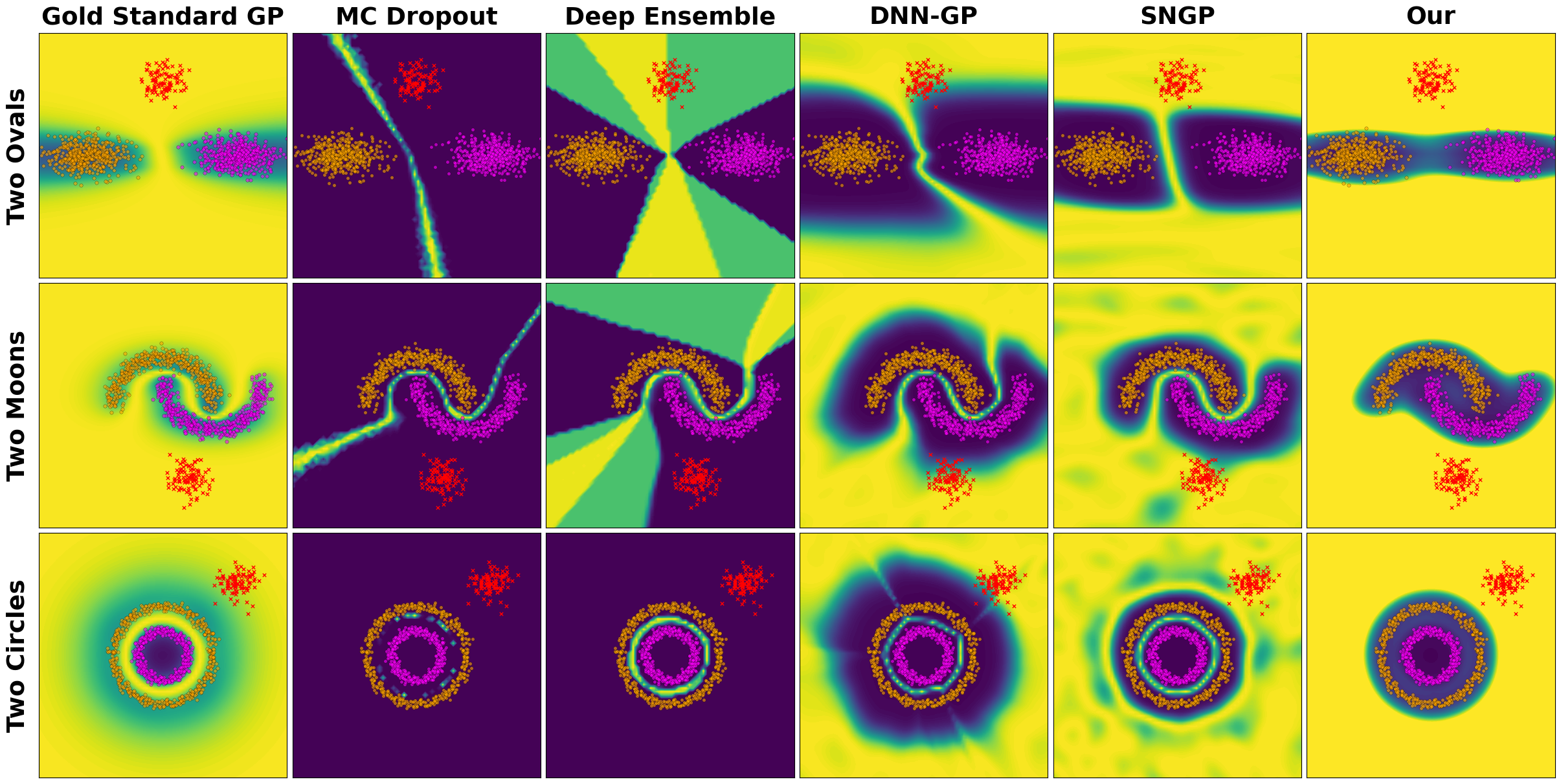}
    \caption{Predictive uncertainty for a Gaussian Process (GP) and neural-network baselines on three toy datasets. \textcolor{orange}{\textbf{Orange}}/\textcolor{magenta}{\textbf{Magenta}} points are in-distribution training samples, and \textcolor{red}{\textbf{Red}} points are out-of-distribution samples. Background shading indicates relative uncertainty (brighter = higher, darker = lower). Our method best matches the GP “gold standard”: it stays confident near observed data and becomes uncertain as inputs move away from the training distribution. }
    \label{fig:uncertaity_prediction_toydata}
    \vspace{-2em}
\end{figure*}

\section{Related Works}\label{sec:related_works}
\textbf{Continuous Environment Mapping:} Continuous environment representations have evolved from \textit{Gaussian Process Occupancy Maps} (GPOM) \cite{GPOM}, which provide principled Bayesian spatial reasoning but suffer from cubic computational complexity, limiting its scalability in large-scale or 3D environments. To address these inefficiencies, \textit{Hilbert Maps} \cite{HilbertMaps, 3D_HilbertMaps} leverage kernel approximations with a simple softmax classifier to construct continuous maps in linear time, while \textit{Bayesian Hilbert Maps} (BHMs) \cite{DBLP:conf/corl/SenanayakeR17, HM} further enable uncertainty estimation by maintaining posterior distributions over model parameters through variational inference. Recently, \textit{V-PRISM} \cite{DBLP:conf/iros/WrightZJH24} introduced a multiclass formulation for mapping tabletop scenes, jointly modeling semantic segmentation, and predictive uncertainty. While Hilbert Maps based methods achieve high accuracy in occupancy and semantic prediction, they tend to produce overconfident estimates in sparsely observed or entirely unobserved regions, which undermines reliability during autonomous operation. On the other hand, Bayesian formulations \cite{DBLP:conf/corl/SenanayakeR17, DBLP:conf/iros/WrightZJH24, sptemp} mitigate this issue by modeling uncertainty through posterior variance, but introducing extensive inference and optimization costs again, which can hinder their real-time deployment in large-scale environments.

\textbf{Uncertainty Estimation in Neural Networks:}
Quantifying predictive uncertainty is vital for the safe deployment of learning-based systems in real-world scenarios. \textit{Spectral-normalized Neural Gaussian Processes} (SNGP) \cite{distance_awareness} encourages predictive uncertainty to increase as inputs move away from the training data by bounding hidden-layer sensitivity via spectral normalization. \textit{Monte Carlo Dropout} (MC Dropout) \cite{MC_Dropout} interprets stochastic dropout at inference time as an approximate Bayesian inference mechanism, while \textit{Deep Ensembles} \cite{Deep_Ensembles} provide a strong non-Bayesian alternative by aggregating multiple independently trained models to capture predictive ambiguity. However, as illustrated in Figure \ref{fig:uncertaity_prediction_toydata}, many of these methods do not consistently produce reliable uncertainty estimation. In particular, despite explicit regularization for distance awareness, SNGP can still exhibit high-confidence predictions in regions far from the observed data.

\textbf{Out-of-Distribution Detection:} Out-of-Distribution (OOD) detection aims to identify inputs that are different from the training distribution, which is a critical safety concern for deep neural networks (DNNs). Because DNNs are typically trained under a closed-set assumption (test samples follow the same distribution as training data), exposure to distribution shifts in real-world environments can lead to unreliable and overconfident predictions \cite{Mohseni_Pitale_Yadawa_Wang_2020, DBLP:journals/corr/abs-1904-12220}. To address this issue, prior works has explored classification-with-rejection and open-set recognition frameworks, which explicitly model the presence of unknown or ambiguous inputs through the rejection rule \cite{cheng2024unified, DBLP:conf/nips/LiuWOL20}. Other methods have explored looking at diffusion models to detect OOD samples \cite{Graham_2023_CVPR, cheng2025dose3}. In the context of environment mapping, we observe that unobserved spatial regions naturally play a role similar to OOD inputs. Our method explicitly samples noise to represent these unobserved regions and treats them as an additional "uncertain" class during training. This formulation aligns with classification-with-rejection principles, enabling the model to allocate probability mass to out-of-distribution class instead of forcing overconfident assignments to known classes. Furthermore, building on this perspective, we theoretically show that the probability assigned to the "uncertain" class is a monotonic function of a distance-aware uncertainty measure.

\begin{figure*}[t]
\centering
\includegraphics[width=0.95\linewidth]{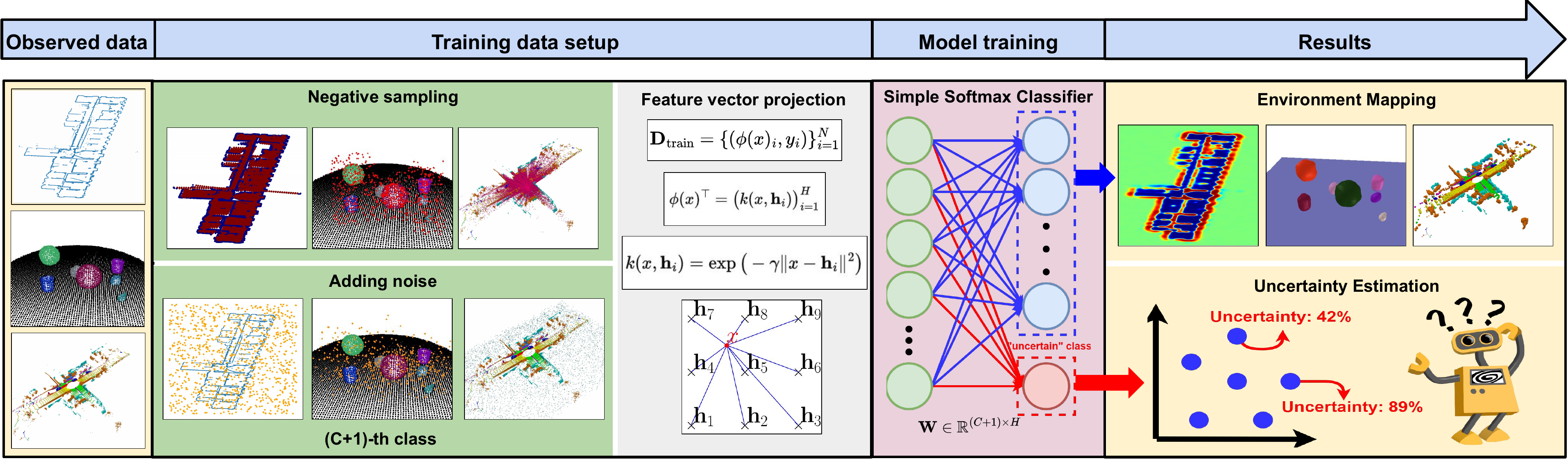}
\caption{Overview of \textbf{ContraMap}. Our method employs a softmax classifier network with an additional output node to jointly represent the environment and estimate uncertainty. Observed data (e.g., LiDAR scan or segmented point cloud) are augmented with negative samples and noise, where the noise is labeled as an additional $(C+1)$-th or "uncertain" class. The augmented data are then projected into a feature vector space using a set of reference points, forming the training dataset $\mathbf{D}_{\mathrm{train}}$. The model is trained on this dataset using a first-order optimization method. Once trained, we can query the model at arbitrary locations in the environment to obtain either environment mapping results or uncertainty estimates via the additional "uncertain" node.}
\vspace{-2em}
\label{fig:method_overview}
\end{figure*}

\section{Preliminaries}\label{sec:2}
Modern robotic mapping methods represent spatial structure as a continuous function rather than a fixed grid, allowing occupancy or semantic information to be queried at any location in space.
Given training samples $\mathbf{X} = {(x_i, y_i)}_{i=1}^{N}$, where $x_i\!\in\!\mathbb{R}^d$ is a spatial coordinate and $y_i$ denotes occupancy or class labels, these models predict the probability of each label using kernel features that encode local spatial correlations.

Each point is projected into a feature vector
\begin{align}
\phi(x)^\top = [k(x,\mathbf{h}_1),\, k(x,\mathbf{h}_2), \ldots,\, k(x,\mathbf{h}_H)],
\end{align}
where $\{\mathbf{h}_i\}_{i=1}^H$ are reference points (kernel centres) and $k(\cdot,\cdot)$ is typically a Gaussian kernel
\begin{align}
k(x,\mathbf{h}) = \exp(-\gamma \|x - \mathbf{h}\|^2),
\end{align}
with bandwidth $\gamma$ controlling the spatial smoothness.
This mapping allows simple linear models on $\phi(x)$ to represent complex, non-linear structures, making kernel-based methods an efficient and interpretable foundation for continuous occupancy and semantic mapping.

\subsection{Bayesian Kernel-based Mapping}

A Bayesian view treats the model weights as random variables, allowing both the mean prediction and its uncertainty to be inferred.
For binary occupancy, the likelihood is
\begin{align}
P(y=1 \mid x, \mathbf{w}) = \sigma(\mathbf{w}^\top \phi(x)),
\end{align}
where $\mathbf{w}$ is given a Gaussian prior $p(\mathbf{w}) = \mathcal{N}(\mu_0, \Sigma_0)$.
The predictive distribution marginalises over the posterior
\begin{align}
P(y=1 \mid x_*, \mathbf{X}) =
\int \sigma(\mathbf{w}^\top \phi(x_*))\, p(\mathbf{w}\mid \mathbf{X})\, d\mathbf{w},
\end{align}
providing both occupancy probability and predictive variance.

For multi-class segmentation, the formulation extends to a softmax likelihood with weight matrix $\mathbf{W}\in\mathbb{R}^{C\times H}$, enabling simultaneous reasoning over multiple object categories.
In both cases, the posterior covariance $\Sigma$ captures spatial correlations and confidence but requires repeated inversion of an $H\times H$ matrix.
This operation scales as $\mathcal{O}(H^3)$, making Bayesian inference computationally expensive for dense or large-scale maps.

\subsection{Motivation}

While Bayesian kernel-based mapping offers a principled measure of uncertainty, its cubic cost prevents real-time deployment in robotic settings where both scale and responsiveness are critical.
Moreover, parameter uncertainty does not always correspond directly to spatial uncertainty, especially in partially observed or occluded regions.

These challenges motivate \textbf{ContraMap}, which preserves the efficiency and spatial continuity of kernel-based models but learns uncertainty explicitly as a predictive quantity.
By introducing an additional ``uncertain'' class and training with contrastive supervision between observed and synthetic noise samples, ContraMap enables joint prediction and uncertainty estimation without Bayesian inference, achieving real-time uncertainty-aware mapping.
\section{Method Overview}\label{sec:3}

Bayesian extensions of Hilbert Maps (HMs) \cite{DBLP:conf/corl/SenanayakeR17, DBLP:conf/iros/WrightZJH24} provide a principled means of modeling uncertainty, but their reliance on covariance matrix inversion leads to cubic scaling with the number of hinge points, making them unsuitable for real-time deployment. Our goal is to design a lightweight alternative that both preserves the efficiency of HMs and captures uncertainty.

\subsection{Noise Contrastive Estimation to Uncertainty Modeling}

Noise Contrastive Estimation (NCE) \cite{DBLP:journals/jmlr/GutmannH12,pmlr-v9-gutmann10a} offers a useful perspective: model parameters can be estimated by training a classifier to distinguish data samples $x \in \mathbf{X}$ drawn from $p_{\mathrm{InD}}(\cdot|\theta)$ against noise samples $\tilde{x} \in \mathbf{\tilde{X}}$ drawn from $p_{\mathrm{noise}}(\cdot)$. The objective is to maximize
\begin{align}
    J(\theta) &= \frac{1}{|\mathbf{X}|}\sum_{x \in \mathbf{X}} \ln[h(x;\theta)] 
    + \frac{1}{|\mathbf{\tilde{X}}|}\sum_{\tilde{x} \in \mathbf{\tilde{X}}} \ln[1-h(\tilde{x};\theta)], \label{eq:nce_obj}
\end{align}
where $h(u;\theta)$ is the posterior probability that $u$ originates from the data distribution. Importantly, this objective is equivalent to the log-likelihood of a logistic regression classifier discriminating between real and noise samples. Thus, NCE establishes a direct connection between density estimation and supervised classification. We leverage this connection to reinterpret uncertainty modeling: rather than inferring full Bayesian posteriors over parameters, we treat ambiguous or unseen regions as “noise” and explicitly introduce them as a separate prediction class.

\subsection{Efficient Hilbert Maps with an Uncertainty Class}
% \subsection{Hilbert Maps}
Building on HMs, which use logistic regression to model occupancy, we extend the sigmoid formulation to a softmax classifier model that includes an additional class for uncertainty. Here, we assume that we have a segmented point cloud and seek to assign each segmentation mask a label, then, learn a continuous representation encapsulating this. Concretely, the model is parameterized by a weight matrix $\mathbf{W} \in \mathbb{R}^{(C+1)\times H}$, where $C$ is the number of standard classes ($C = 2$ for 2D mapping, i.e., occupied/free). To train $\mathbf{W}$, we construct an augmented dataset:
\begin{align}
    \mathbf{D} = \{(x_i, y_i)\}_{i=1}^n \cup \{(x^*_j, y^*_j = C+1)\}_{j=1}^m,
\end{align}
where $(x_i,y_i)$ are the original samples and $\{(x^*_j,y^*_j=C+1)\}$ are noise points randomly sampled from the environment, labeled as the $(C+1)$-th ``uncertain'' class. 

Training then reduces to optimizing the standard cross-entropy loss over $\mathbf{D}$, which can be done efficiently using gradient descent. The resulting model outputs occupancy probabilities for known classes while the $(C+1)$-th node provides an explicit estimate of uncertainty. This design retains the scalability and efficiency of HMs while avoiding the cubic computational burden of Bayesian variants.

\section{From Noise to an Uncertainty Indicator}
\label{sec:noise_to_uncertainty}

Our classifier is trained with an explicit extra label $(C\!+\!1)$ representing \emph{noise} or \emph{none-of-the-above} inputs. At test time, this allows the model to either: (i) explain a query using one of the $C$ in-distribution classes, or (ii) allocate probability mass to the uncertainty class when the query lies away from the support of the observed data. In this sense, the additional output should be interpreted primarily as a learned distance-aware OOD score over continuous space.

The goal of this section is therefore not to claim calibrated uncertainty in a strict probabilistic sense, but to justify why the softmax probability on the $(C\!+\!1)$-th node, $p_{C+1}(x)$, provides a meaningful uncertainty \emph{ordering}. In particular, we show that under a simple mixture-model view, $p_{C+1}(x)$ is a monotonic function of a distance-aware uncertainty surrogate.

\noindent\textbf{Key idea:}
If we train on a mixture of (a) in-distribution data and (b) broadly spread noise labeled as $(C\!+\!1)$, then a well-trained softmax classifier approximates the Bayes posterior for the noise component. In this case, $p_{C+1}(x)$ becomes (approximately) the probability that $x$ came from the noise process rather than the in-distribution process. Since in-distribution density typically \emph{decays} as we move away from the data manifold, this posterior naturally \emph{increases} with distance-to-data, which is also how many uncertainty measures behave.

\noindent\textbf{Distance to the in-distribution:}
Let $\mathbf{X}_{\mathrm{InD}}$ be the in-distribution training set. We quantify how far a test input $x$ lies from the in-distribution support via the expected distance
\begin{align}
d(x) \;=\; \mathbb{E}_{x' \sim p_{\mathrm{InD}}}\!\bigl[\|x - x'\|_\mathbf{X}\bigr],
\label{eq:distance_def_intuitive}
\end{align}
where $\|\cdot\|_\mathbf{X}$ is the distance metric induced by the input space.

\noindent\textbf{A simple generative picture:}
To interpret the uncertainty node, we consider a stylised mixture model in which training examples are drawn from:
\begin{align}
p(x,y) \;=\; \alpha_{\mathrm{InD}}\, p_{\mathrm{InD}}(x,y) \;+\; \alpha_{\mathrm{noise}}\, p_{\mathrm{noise}}(x, y=C+1),
\label{eq:mix_model_intuitive}
\end{align}
and the noise inputs are approximately uniform in the input space, i.e., $p_{\mathrm{noise}}(x)\approx n_0$.
Also assume the in-distribution marginal density decreases as we move away from the data, meaning $p_{\mathrm{InD}}(x) \approx g(d(x))$,
with $g$ positive and non-increasing.

\noindent\textbf{What does the $(C\!+\!1)$ softmax learn?}
For a sufficiently well-trained softmax classifier, the $(C\!+\!1)$ output approximates the Bayes posterior probability \cite{GoodBengCour16} that $x$ was generated by the noise component:
\begin{align}
p_{C+1}(x)
\;\approx\;
\frac{\alpha_{\mathrm{noise}}\, p_{\mathrm{noise}}(x)}
{\alpha_{\mathrm{InD}}\, p_{\mathrm{InD}}(x) + \alpha_{\mathrm{noise}}\, p_{\mathrm{noise}}(x)}.
\label{eq:bayes_posterior_intuitive}
\end{align}
Plugging in $p_{\mathrm{noise}}(x)\approx n_0$ and $p_{\mathrm{InD}}(x)\approx g(d(x))$ gives
\begin{align}
p_{C+1}(x)
\;\approx\;
\frac{\alpha_{\mathrm{noise}}\, n_0}{\alpha_{\mathrm{InD}}\, g(d(x)) + \alpha_{\mathrm{noise}}\, n_0}.
\label{eq:noise_post_as_distance}
\end{align}
Because $g(\cdot)$ is non-increasing, the denominator shrinks as $d(x)$ grows, so
\emph{$p_{C+1}(x)$ is non-decreasing in the distance-to-data $d(x)$.}
In other words, the farther we move from the training support, the more the classifier assigns a higher probability to the noise class.

\noindent\textbf{Connecting to an uncertainty measure:}
Many standard uncertainty measures are also distance-aware, i.e., there exists a non-decreasing function $h$ such that
\begin{align}
U(x) = h(d(x)).
\label{eq:distance_aware_uncertainty}
\end{align}
This is known to hold, for example, for Gaussian Processes with RBF kernels where predictive variance increases with distance from the training data~\cite{distance_awareness}. Combining \eqref{eq:noise_post_as_distance} and \eqref{eq:distance_aware_uncertainty}, we obtain an (input-dependent) monotone relationship:
\begin{align}
p_{C+1}(x) \;\approx\; \varphi\!\left(U(x)\right),
\label{eq:justification_intuitive}
\end{align}
for some non-decreasing function $\varphi$ (implicitly given by composing the distance-to-noise-posterior mapping with the inverse of $h$).

Under these assumptions, the output probability of the $(C+1)$-th class is a monotonic transformation of an input-dependent uncertainty measure. This does not imply that $p_{C+1}(x)$ is a calibrated uncertainty estimate in the strict probabilistic sense; rather, it shows that the score preserves the relative ordering of uncertainty and can therefore serve as a useful spatial uncertainty indicator.
\section{Experiments}\label{sec:5}
To demonstrate the effectiveness of the proposed method, we conduct a comprehensive set of experiments evaluating predictive uncertainty, mapping accuracy, and scalability across both 2D and 3D robotic environments. Specifically, we provide qualitative uncertainty analysis and quantitative comparisons against Hilbert Maps, Bayesian Hilbert Maps, V-PRISM, and 3D Hilbert Maps, as well as an evaluation on hinge point count and scalability. All experiments are run on a NVIDIA Tesla P100 GPU.
% ======================================================
\subsection{Qualitative evaluation of predictive uncertainty}
To qualitatively evaluate uncertainty estimation behavior, we first conducted experiments on three standard toy classification datasets: Two Ovals, Two Moons, and Two Circles. These datasets feature increasingly complex decision boundaries and are commonly used to analyze uncertainty. Following the setup in prior work on uncertainty estimation \cite{distance_awareness}, we compare our approach against a Gold Standard Gaussian Process (GP), MC Dropout \cite{MC_Dropout}, Deep Ensembles \cite{Deep_Ensembles}, DNN-GP and SNGP \cite{distance_awareness}. As these methods are not designed for environment mapping, the comparison focuses exclusively on uncertainty estimation behavior, rather than quantitative classification accuracy. Uncertainty in our method is derived directly through the $(C+1)$-th node in the output layer, while the compared methods infer uncertainty implicitly from their predictive distributions. As shown in Figure \ref{fig:uncertaity_prediction_toydata}, the Gold Standard GP exhibits low uncertainty in the vicinity of training samples and increasing uncertainty as predictions move farther from the observed data. Among the compared methods, our approach most closely matches this behavior across all three datasets. In our results, uncertainty remains low near the data manifold and increases smoothly in regions that are sparsely sampled or entirely unobserved, including areas containing out-of-distribution points.

This behavior of our model stems from the structure inherited from Hilbert Maps \cite{HilbertMaps}. By projecting inputs into a high-dimensional space using hinge points, the model can effectively separate complex class geometries. In addition, we introduce uniformly sampled noise labeled as an explicit "uncertain" class during training. These samples act as negative evidence across the input space, enabling the model to identify regions that lack sufficient training evidence. Consequently, unlike methods that derive uncertainty implicitly from the predictive class distribution (e.g., entropy or variance), uncertainty in our approach is governed by distance from the training distribution rather than ambiguity between in-distribution classes, and no elevated uncertainty is observed along boundaries between well-separated classes.

% ======================================================
\subsection{Compare to HMs and BHMs in occupancy mapping}
This experiment empirically evaluates the proposed method in terms of accuracy, training time, and inference time; and compares it with Hilbert Maps (HMs) \cite{HilbertMaps} and Bayesian Hilbert Maps (BHMs) \cite{DBLP:conf/corl/SenanayakeR17} in occupancy mapping. We also demonstrate the ability of our approach to capture uncertainty by extracting outputs for each class in the training dataset $\mathbf{D}$, particularly the $(C+1)$-th class.
\subsubsection{Settings}
Experiments are conducted on multiple datasets from the Radish repository \cite{Radish}, each is split into training and testing sets with a 9:1 ratio. HMs and BHMs are trained using the observed training data $\mathbf{X}$ derived directly from Radish datasets, while our softmax model is trained on an augmented dataset $\mathbf{D}$ formed by combining $\mathbf{X}$ with randomly sampled noise from the environment.

In this occupancy mapping task, points in $\mathbf{X}$ are labeled as $c=1$ (occupied) or $c=0$ (unoccupied). To construct $\mathbf{D}$, we generate a noise set $\mathbf{\tilde{X}}$ with $|\mathbf{\tilde{X}}|=|\mathbf{X}|$ by uniformly sampling points across the landscape and labeling them as $c=2$, representing an additional uncertainty class. Thus, $\mathbf{D}=\mathbf{X}\cup\mathbf{\tilde{X}}$ contains three classes, and the softmax model has three output nodes. For feature computation, the hinge points $\{\mathbf{h}_i\}_{i=1}^H$ are chosen evenly across the landscape, with $H$ determined by the map size. 

Performance is evaluated using the Area Under the Receiver Operating Characteristic Curve (AUC) \cite{AUC}, together with average training and inference times. The AUC is computed from predictions on the testing set. For inference, evenly spaced query points are generated to cover each map; these points are batch-processed by each model to produce occupancy maps, and inference times are recorded.

\subsubsection{Results Analysis}
\begin{table*}[t]
\centering
\caption{The performance of our method in comparison to Hilbert Map \cite{HilbertMaps} and Bayesian Hilbert Map \cite{DBLP:conf/corl/SenanayakeR17} on the Radish datasets \cite{Radish}.}\label{tab:1}
\begin{tabular}{|cc|c|c|c|}
\hline
\multicolumn{2}{|c|}{\textbf{Datasets}}                                            & \textbf{Hilbert Maps \cite{HilbertMaps}} & \textbf{Bayesian Hilbert Maps \cite{DBLP:conf/corl/SenanayakeR17}} & \textbf{Our}                   \\ \hline
% \multicolumn{1}{|c|}{\multirow{3}{*}{\makecell{\textbf{Aces} \\ (\#features = 3900)}}}       & AUC $\uparrow$                 & 0.9075 ($\pm$ 0.0026)          & \textbf{0.9121 ($\pm$ 0.0025)} & 0.9116 ($\pm$ 0.0024)          \\
% \multicolumn{1}{|c|}{}                                                             & Training time (s) $\downarrow$ & 1.7648 ($\pm$ 0.2720)          & 122.3577 ($\pm$ 5.4550)        & \textbf{1.5441 ($\pm$ 0.0261)} \\
% \multicolumn{1}{|c|}{}                                                             & Inference time (s) $\downarrow$  & \textbf{0.1965 ($\pm$ 0.0183)} & 40.9828 ($\pm$ 0.0867)         & 0.2075 ($\pm$ 0.0147)          \\ \hline
\multicolumn{1}{|c|}{\multirow{3}{*}{\makecell{\textbf{Belgioioso} \\ (\#features = 8400)}}} & AUC $\uparrow$                 & \textbf{0.9904 ($\pm$ 0.0003)} & 0.9901 ($\pm$ 0.0003)          & 0.9901 ($\pm$ 0.0003)          \\
\multicolumn{1}{|c|}{}                                                             & Training time (s) $\downarrow$ & \textbf{2.2609 ($\pm$ 0.0279)} & 722.6140 ($\pm$ 2.5651)        & 3.2367 ($\pm$ 0.0231)          \\
\multicolumn{1}{|c|}{}                                                             & Inference time (s) $\downarrow$  & \textbf{0.4258 ($\pm$ 0.0014)} & 231.3024 ($\pm$ 0.0746)        & 0.4681 ($\pm$ 0.0022)          \\ \hline
\multicolumn{1}{|c|}{\multirow{3}{*}{\makecell{\textbf{Edmonton}\\ (\#features = 8800)}}}    & AUC $\uparrow$                 & 0.9754 ($\pm$ 0.0005)          & 0.9734 ($\pm$ 0.0005)          & \textbf{0.9767 ($\pm$ 0.0004)} \\
\multicolumn{1}{|c|}{}                                                             & Training time (s) $\downarrow$ & \textbf{1.8345 ($\pm$ 0.0371)} & 789.5725 ($\pm$ 1.7593)        & 2.2733 ($\pm$ 0.0245)          \\
\multicolumn{1}{|c|}{}                                                             & Inference time (s) $\downarrow$  & \textbf{0.6997 ($\pm$ 0.0016)} & 405.7368 ($\pm$ 0.0842)        & 0.7630 ($\pm$ 0.0011)          \\ \hline
\multicolumn{1}{|c|}{\multirow{3}{*}{\makecell{\textbf{Fhw} \\ (\#features = 5940)}}}        & AUC $\uparrow$                 & \textbf{0.9608 ($\pm$ 0.0014)} & 0.9583 ($\pm$ 0.0014)          & 0.9603 ($\pm$ 0.0013)          \\
\multicolumn{1}{|c|}{}                                                             & Training time (s) $\downarrow$ & \textbf{1.6850 ($\pm$ 0.0467)} & 326.7038 ($\pm$ 0.8631)        & 1.9292 ($\pm$ 0.0125)          \\
\multicolumn{1}{|c|}{}                                                             & Inference time (s) $\downarrow$  & \textbf{0.2186 ($\pm$ 0.0127)} & 75.1121 ($\pm$ 0.0267)         & 0.2358 ($\pm$ 0.0008)          \\ \hline
\multicolumn{1}{|c|}{\multirow{3}{*}{\makecell{\textbf{Intel} \\ (\#features = 5600)}}}      & AUC $\uparrow$                 & 0.9644 ($\pm$ 0.0009)          & \textbf{0.9688 ($\pm$ 0.0007)} & 0.9631 ($\pm$ 0.0008)          \\
\multicolumn{1}{|c|}{}                                                             & Training time (s) $\downarrow$ & \textbf{1.9923 ($\pm$ 0.2506)} & 288.2842 ($\pm$ 1.7646)        & 2.5446 ($\pm$ 0.0206)          \\
\multicolumn{1}{|c|}{}                                                             & Inference time (s) $\downarrow$  & \textbf{0.1977 ($\pm$ 0.0030)} & 65.4582 ($\pm$ 0.1118)         & 0.2164 ($\pm$ 0.0049)          \\ \hline
\multicolumn{1}{|c|}{\multirow{3}{*}{\makecell{\textbf{Mexico} \\ (\#features = 7500)}}}     & AUC $\uparrow$                 & 0.9726 ($\pm$ 0.0006)          & 0.9709 ($\pm$ 0.0007)          & \textbf{0.9731 ($\pm$ 0.0006)} \\
\multicolumn{1}{|c|}{}                                                             & Training time (s) $\downarrow$ & \textbf{1.7175 ($\pm$ 0.0415)} & 566.5886 ($\pm$ 1.2325)        & 2.0832 ($\pm$ 0.0271)          \\
\multicolumn{1}{|c|}{}                                                             & Inference time (s) $\downarrow$  & \textbf{0.5305 ($\pm$ 0.0042)} & 251.3564 ($\pm$ 0.1505)        & 0.5786 ($\pm$ 0.0034)          \\ \hline
% \multicolumn{1}{|c|}{\multirow{3}{*}{\makecell{\textbf{Orebro} \\ (\#features = 5600)}}}     & AUC $\uparrow$                 & 0.9374 ($\pm$ 0.0015)          & 0.9393 ($\pm$ 0.0013)          & \textbf{0.9413 ($\pm$ 0.0014)} \\
% \multicolumn{1}{|c|}{}                                                             & Training time (s) $\downarrow$ & \textbf{1.6476 ($\pm$ 0.0457)} & 280.0770 ($\pm$ 0.6066)        & 1.8424 ($\pm$ 0.0222)          \\
% \multicolumn{1}{|c|}{}                                                             & Inference time (s) $\downarrow$  & \textbf{0.2009 ($\pm$ 0.0106)} & 65.2806 ($\pm$ 0.0381)         & 0.2145 ($\pm$ 0.0007)          \\ \hline
% \multicolumn{1}{|c|}{\multirow{3}{*}{\makecell{\textbf{Seattle} \\ (\#features = 2100)}}}    & AUC $\uparrow$                 & 0.9621 ($\pm$ 0.0011)          & \textbf{0.9684 ($\pm$ 0.0011)} & 0.9622 ($\pm$ 0.0009)          \\
% \multicolumn{1}{|c|}{}                                                             & Training time (s) $\downarrow$ & 1.6513 ($\pm$ 0.0206)          & 18.2961 ($\pm$ 0.1005)         & \textbf{1.5525 ($\pm$ 0.0343)} \\
% \multicolumn{1}{|c|}{}                                                             & Inference time (s) $\downarrow$  & \textbf{0.0587 ($\pm$ 0.0062)} & 6.5392 ($\pm$ 0.0354)          & 0.0737 ($\pm$ 0.0034)          \\ \hline
\end{tabular}
\vspace{-2em}
\end{table*}

\begin{table}[t]
    \centering
    \caption{Quantitative comparison against 3D Hilbert Maps (HMs) on the SemanticKITTI dataset.
    Experiments were conducted on the initial scan of 10 distinct sequences. \textbf{Bold} indicates the statistically better result ($p < 0.05$).}
    \label{tab:semantickitti}
    % \vspace{0.2cm}
    \begin{tabular}{|c|c|c|}
        \hline
        \textbf{Metric} & \textbf{HMs} & \textbf{Our} \\
        \hline
        mIoU $\uparrow$ & 
            0.8096 $\pm$ 0.0515 & 
            \textbf{0.8412 $\pm$ 0.0394} \\

        Training Time (s) $\downarrow$ & 
            \textbf{32.6278 $\pm$ 1.5994} & 
            35.9025 $\pm$ 1.7746 \\
        
        Inference Time (s) $\downarrow$ & 
            \textbf{203.6160 $\pm$ 8.2965} & 
            203.7221 $\pm$ 8.3205 \\

        \hline
    \end{tabular}
    % \vspace{0.1cm}
    
    % \footnotesize{($\uparrow$) higher is better, ($\downarrow$) lower is better.}
\end{table}
% -----------------------------------------------

% SceneNet
% -----------------------------------------------

\begin{table}[t]
    \centering
    \caption{Quantitative comparison against 3D Hilbert Maps (HMs) on the SceneNet dataset.
    The evaluation was performed across 60 distinct indoor scenes (rooms).
    \textbf{Bold} indicates the statistically better result ($p < 0.05$).}
    \label{tab:scenenet}
    % \vspace{0.2cm}
    \begin{tabular}{|c|c|c|}
        \hline
        \textbf{Metric} & \textbf{HMs} & \textbf{Our} \\
        \hline
        mIoU $\uparrow$ & 
            0.6622 $\pm$ 0.1234 & 
            \textbf{0.6694 $\pm$ 0.1231} \\

        Training Time (s) $\downarrow$ & 
            \textbf{32.0929 $\pm$ 22.6771} & 
            35.9777 $\pm$ 25.5202 \\
        
        Inference Time (s) $\downarrow$ & 
            32.2525 $\pm$ 2.6430 & 
            32.2989 $\pm$ 2.6393 \\
            
        \hline
    \end{tabular}
    % \vspace{0.1cm}
    
    % \footnotesize{($\uparrow$) higher is better, ($\downarrow$) lower is better.}
\end{table}

\begin{figure}[t]
    \centering
    \resizebox{\columnwidth}{!}{
    % --- Subfigure (a) AUC ---
    \fbox{
    \begin{subfigure}{0.32\columnwidth}
        \centering
        \includegraphics[width=\linewidth]{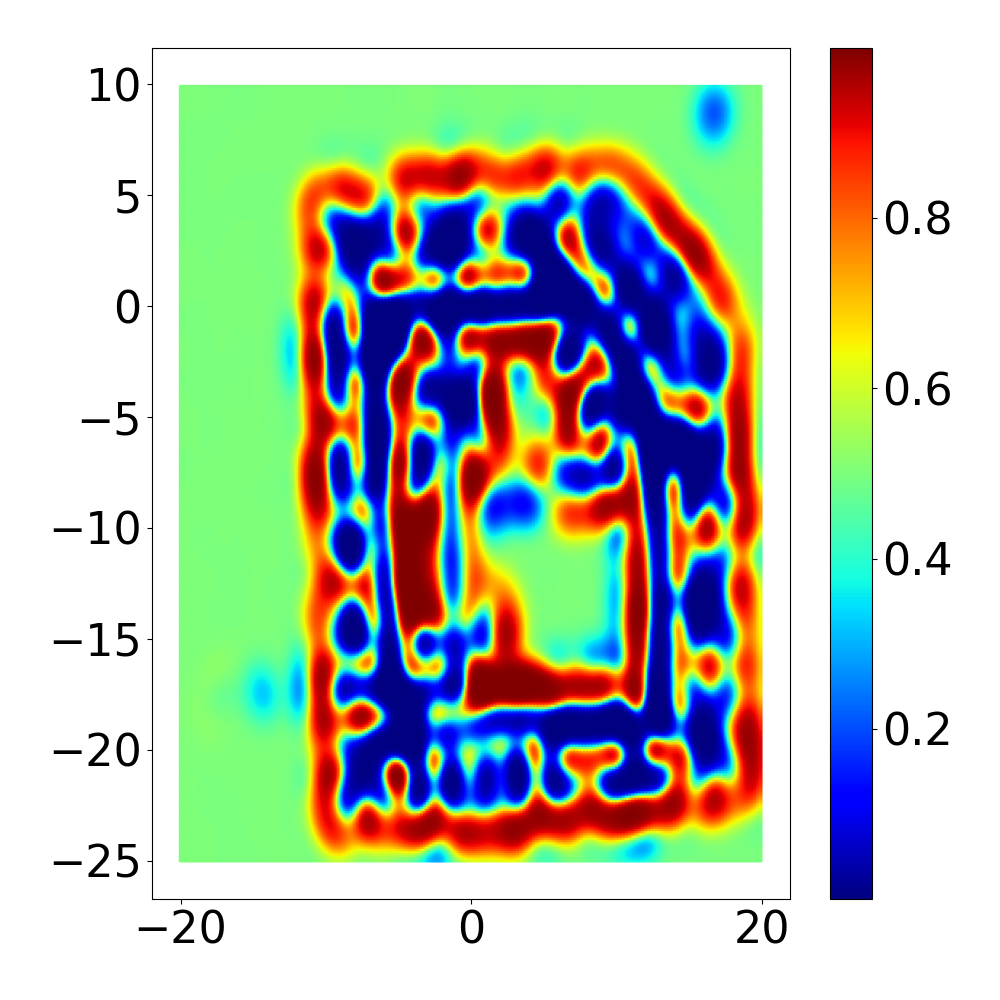}
        \caption{Our}
        \label{fig:hm_probs_our}
    \end{subfigure}
    % --- Subfigure (b) Training time ---
    \begin{subfigure}{0.32\columnwidth}
        \centering
        \includegraphics[width=\linewidth]{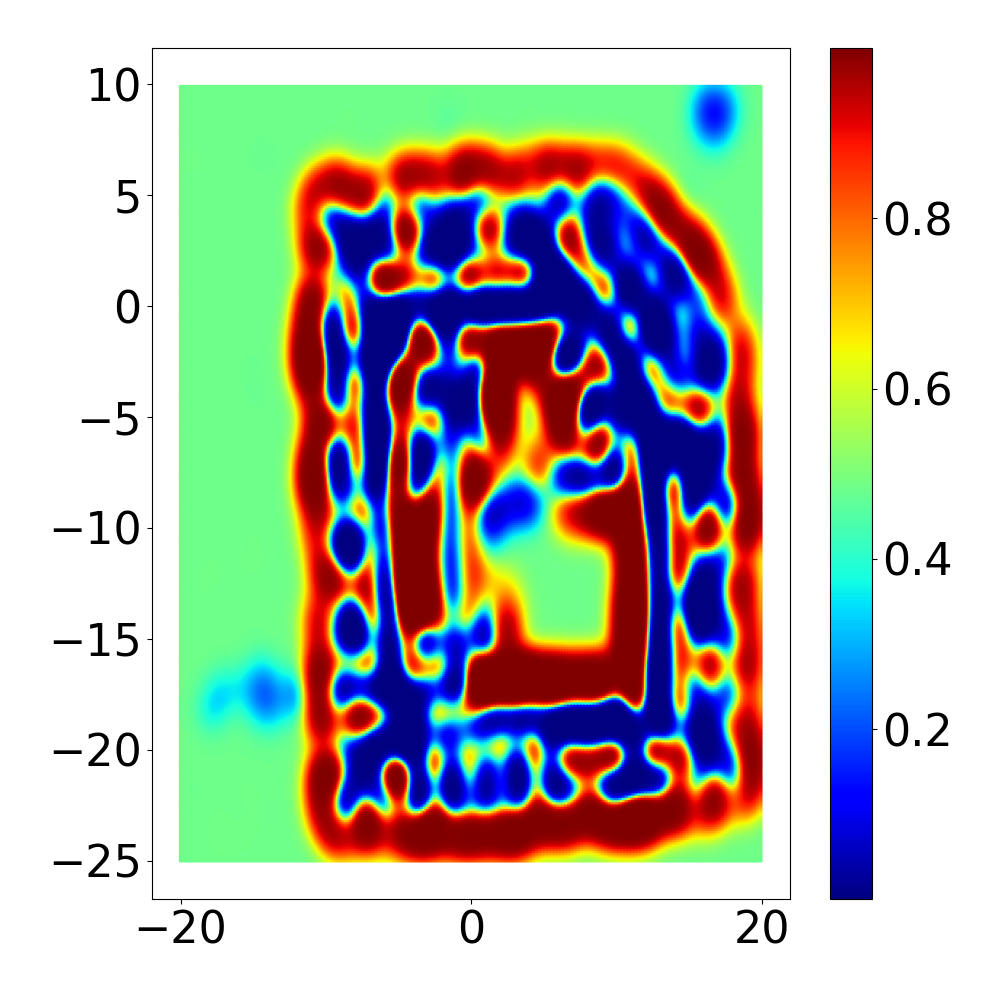}
        \caption{HMs \cite{HilbertMaps}}
        \label{fig:hm_probs_HMs}
    \end{subfigure}
    % --- Subfigure (c) Inference time ---
    \begin{subfigure}{0.32\columnwidth}
        \centering
        \includegraphics[width=\linewidth]{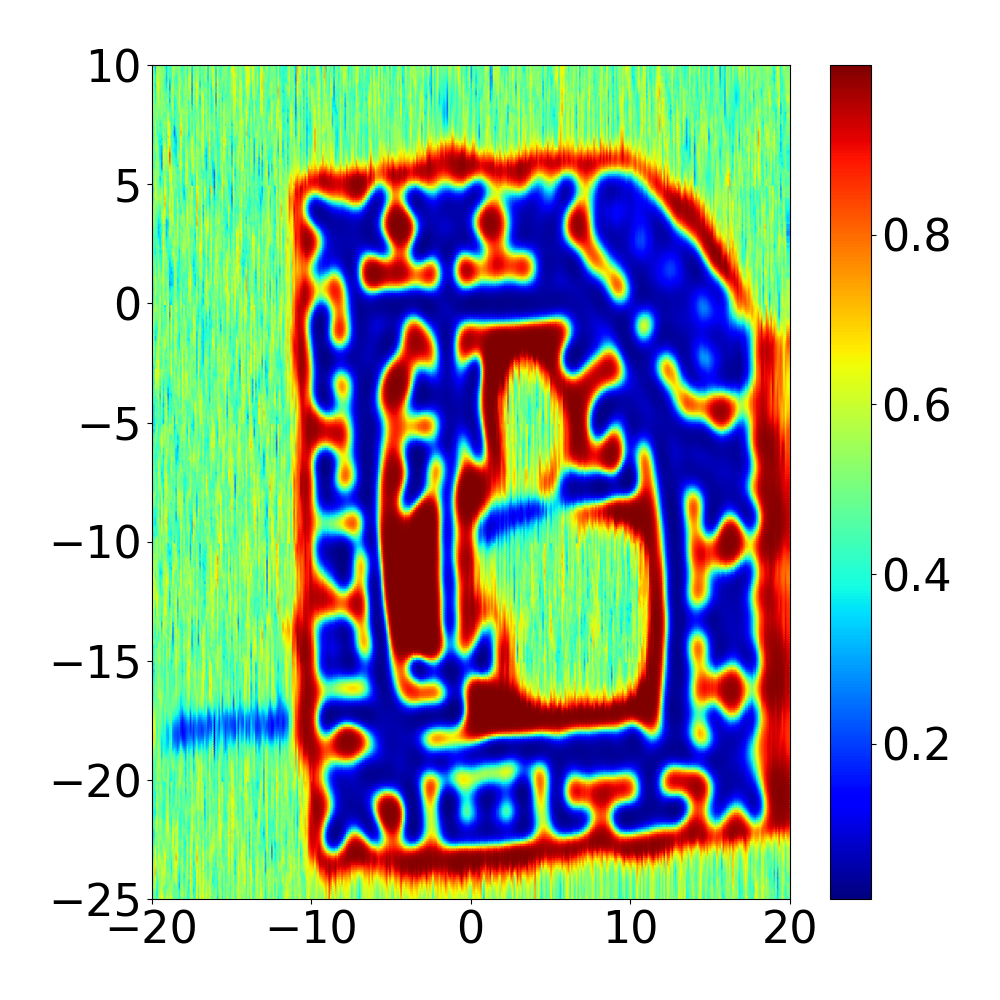}
        \caption{BHMs \cite{DBLP:conf/corl/SenanayakeR17}}
        \label{fig:hm_probs_BHMs}
    \end{subfigure}
    }}
    \caption{Occupancy mapping results of each method on Intel dataset.}
    \label{fig:mapping_results_intel}
\end{figure}

\begin{figure}[t]
    \centering
    % --- Subfigure (a) AUC ---
    \resizebox{\columnwidth}{!}{
    \fbox{
    \begin{subfigure}{0.32\columnwidth}
        \centering
        \includegraphics[width=\linewidth]{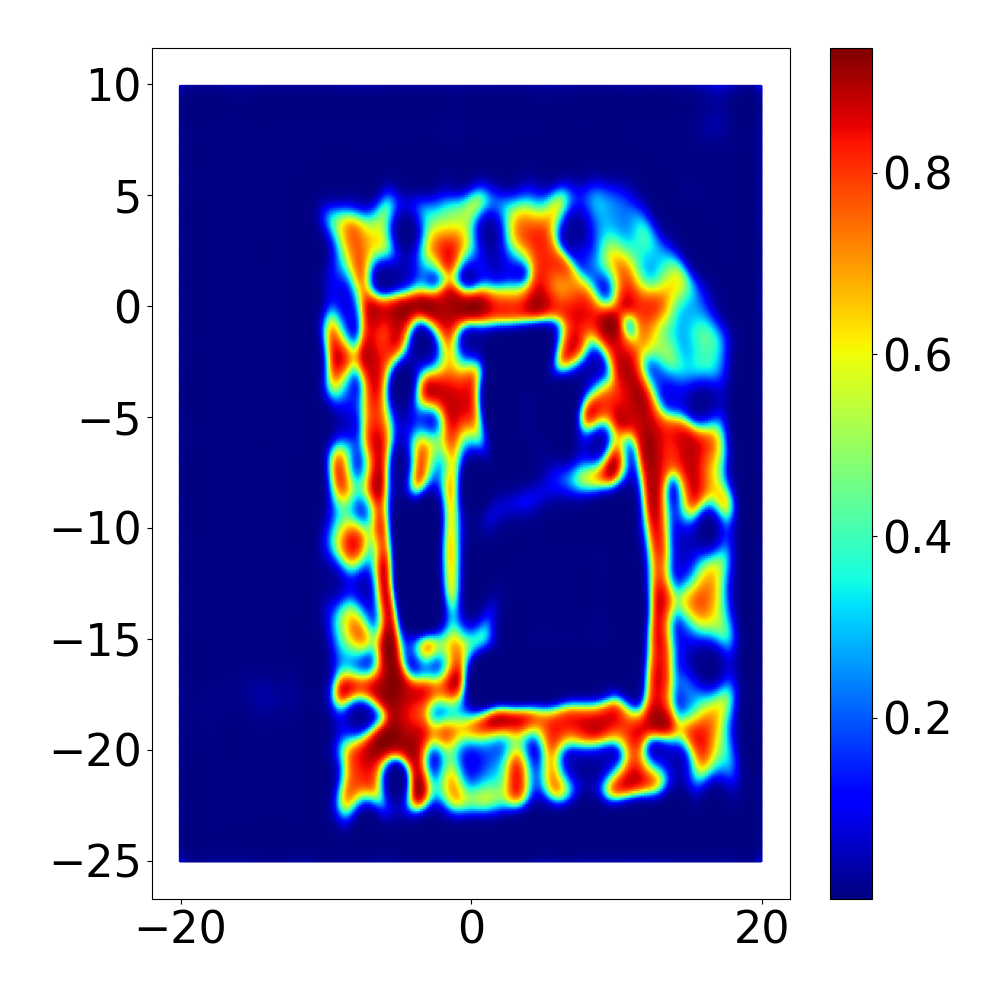}
        \caption{Free Space}
        \label{fig:node0}
    \end{subfigure}
    % --- Subfigure (b) Training time ---
    \begin{subfigure}{0.32\columnwidth}
        \centering
        \includegraphics[width=\linewidth]{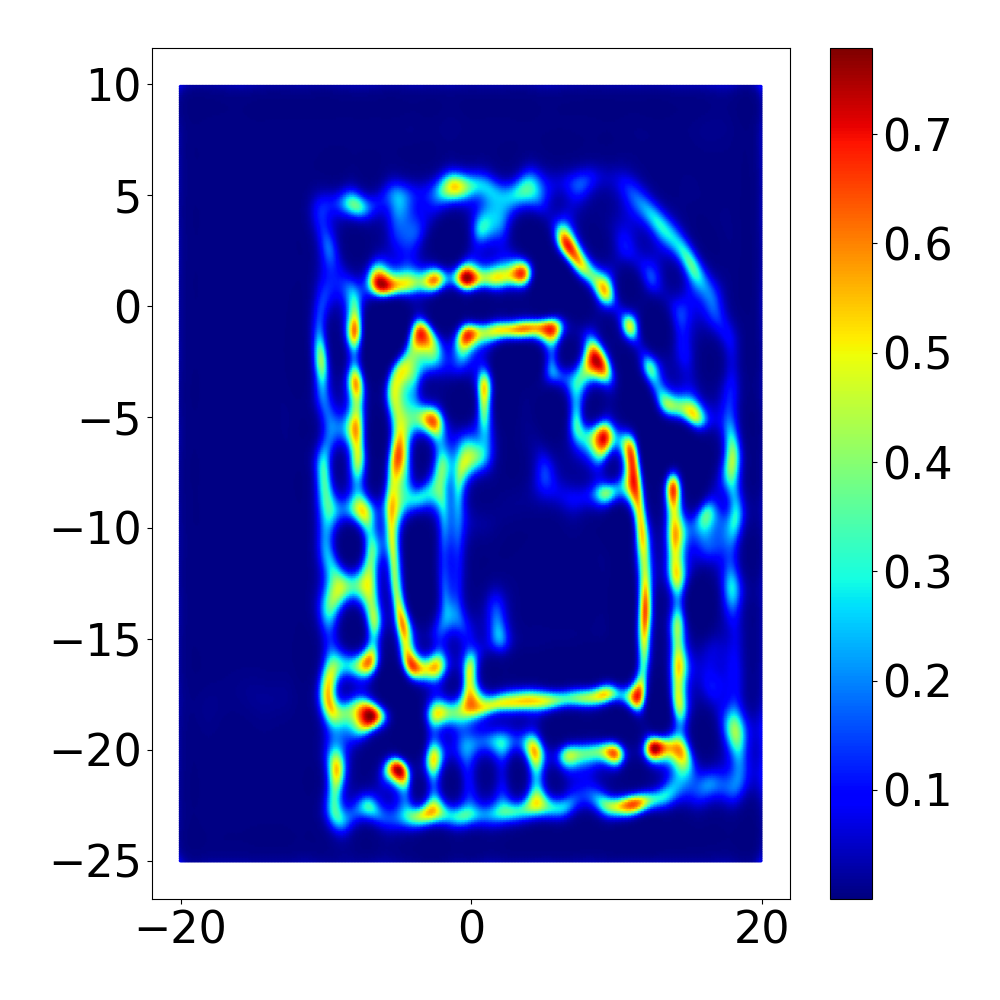}
        \caption{Occupied Space}
        \label{fig:node1}
    \end{subfigure}
    % --- Subfigure (c) Inference time ---
    \begin{subfigure}{0.32\columnwidth}
        \centering
        \includegraphics[width=\linewidth]{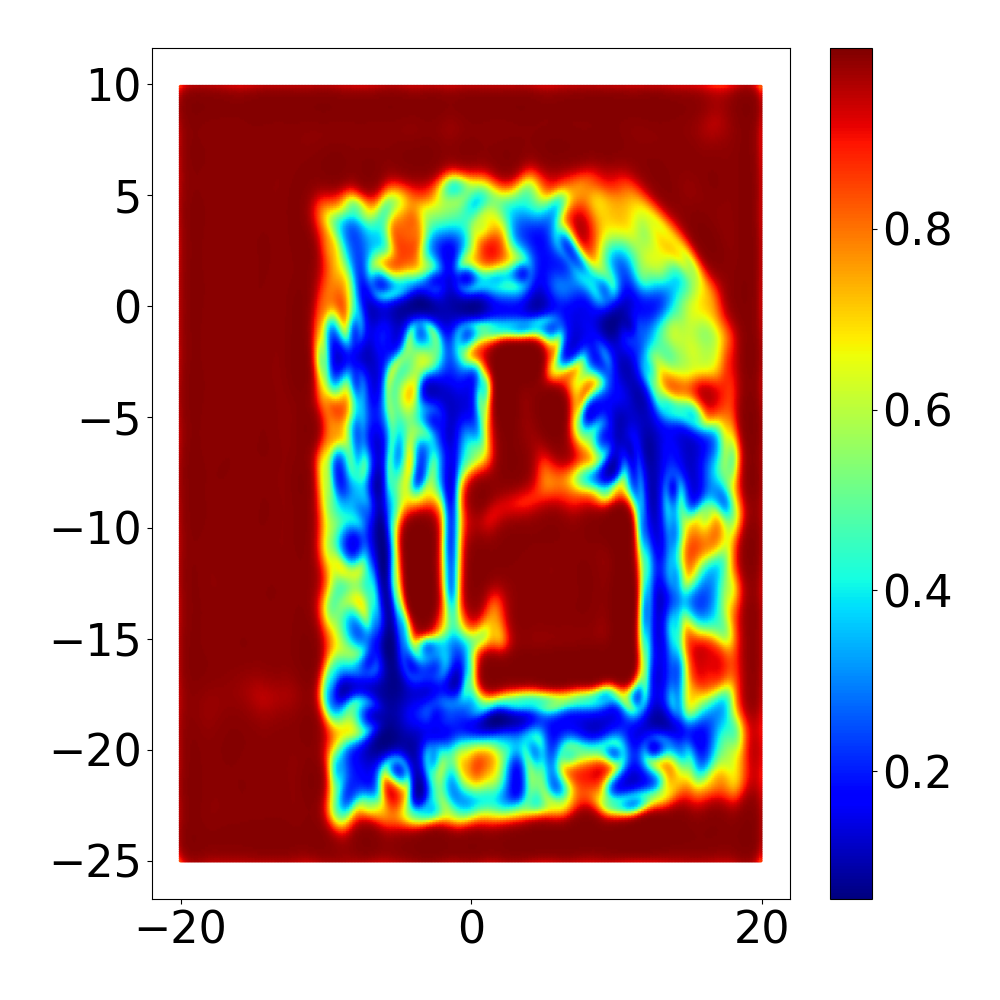}
        \caption{Uncertainty in Space}
        \label{fig:node2}
    \end{subfigure}
    }}
    \caption{Results extracted from each output of the softmax model.}
    \label{fig:softmax_nodes}
\end{figure}

Table \ref{tab:1} shows the accuracy and runtime of each method on different datasets from Radish \cite{Radish}, along with the number of features used for each dataset.

From Table \ref{tab:1}, both our method and HMs require substantially lower training and inference times than BHMs, while achieving comparable mapping accuracy in terms of AUC. This demonstrates the efficiency of parameter learning via gradient descent compared to Bayesian approaches. Since the proposed method and HMs employ simple softmax and logistic regression models, they can be trained effectively using first-order gradient-based optimization, and inference is fast because it only involves forward computation. In terms of runtime, HMs are marginally faster than our method, likely because logistic regression is slightly lighter than softmax model; however, the difference is negligible.
% From Table \ref{tab:1}, we observe that our method and HMs require remarkably lower runtime for both training and inference than BHMs, while the mapping accuracies as assessed by AUC scores between these methods show no significant difference. These results highlight the efficiency of parameters learning via gradient descent technique, in comparison to using Bayesian method. As the proposed method and HMs adopt simple softmax and logistic regression architectures, they can be trained effectively using a first-order gradient-based optimization. Moreover, since the inference process of these models involves only forward computation, the inference phase is even faster. When comparing the runtime between our method and HMs, HMs is marginally faster. This can be explained as the logistic regression model is slightly lighter than the softmax model; nevertheless, this gap turns out to be negligible.

The mapping quality is illustrated in Figure \ref{fig:mapping_results_intel}. Since our method uses a softmax model with three output nodes, the displayed occupancy map is obtained by dividing the output of class $1$ (probability of being occupied) by that of class $0$ (probability of being unoccupied). The results show that our method produces maps with precision comparable to HMs and BHMs.

To further analyze the model, we separately extract the outputs of each node. Figure \ref{fig:softmax_nodes} presents these results on the Intel dataset, showing that each node clearly captures its corresponding class. From Figure \ref{fig:node0}, confidently unoccupied regions are highlighted, while Figure \ref{fig:node1} outlines the occupied areas. Notably, Figure \ref{fig:node2} shows the output of the uncertainty class learned from noise, demonstrating that the additional class effectively models ambiguity. As observed in Figure \ref{fig:node2}, regions confidently classified as class $0$ or $1$ exhibit low uncertainty, whereas unobserved regions, where noise samples are more likely drawn, show high uncertainty. Taken together, the above experiments suggest that our method can perform occupancy mapping accurately with a small runtime, while also be able to explicitly estimate the uncertainty in the environment.
% =====================================================
\subsection{Compare against 3D Hilbert Maps in 3D environments mapping}

\begin{figure}[t]
    \centering
     % --- Column Headers ---
    \begin{subfigure}{0.32\linewidth}
        \centering
        \textbf{Ground Truth}
    \end{subfigure}
    \begin{subfigure}{0.32\linewidth}
        \centering
        \textbf{HMs Results}
    \end{subfigure}
    \begin{subfigure}{0.32\linewidth}
        \centering
        \textbf{Our Results}
    \end{subfigure}
    
 % --- Row 1 ---
    \begin{subfigure}{0.32\linewidth}
        \includegraphics[width=\linewidth]{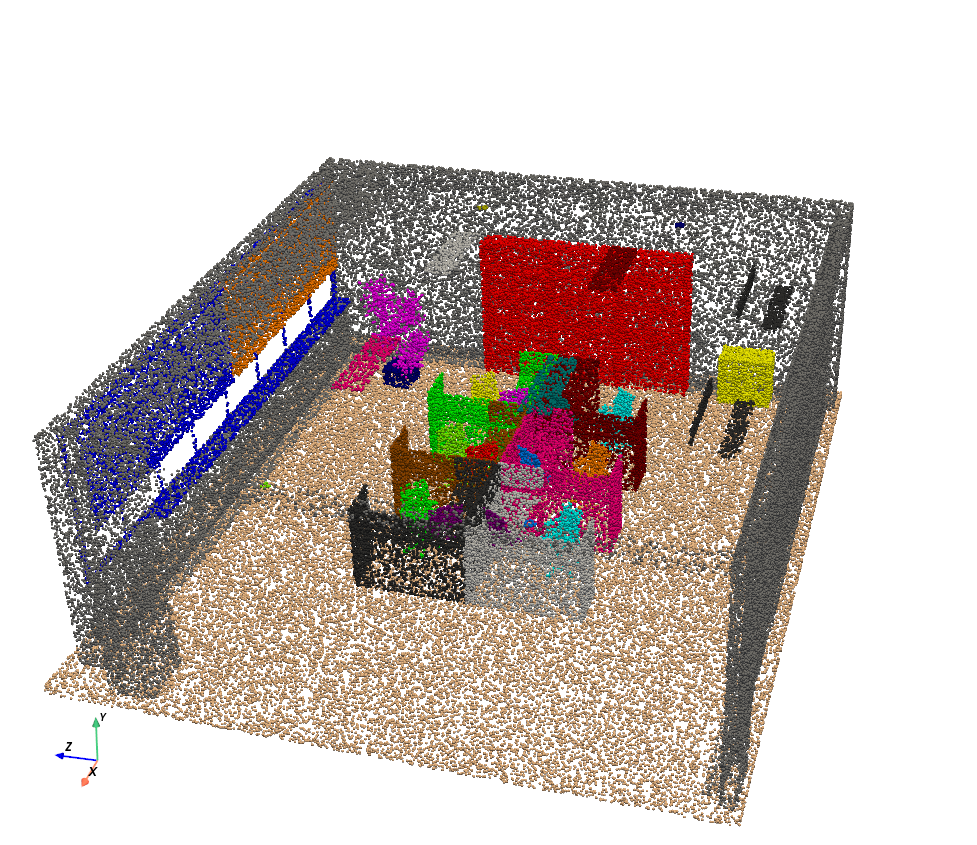}
        % \caption{}
        \label{fig:row1-gt}
    \end{subfigure}
    \begin{subfigure}{0.32\linewidth}
        \includegraphics[width=\linewidth]{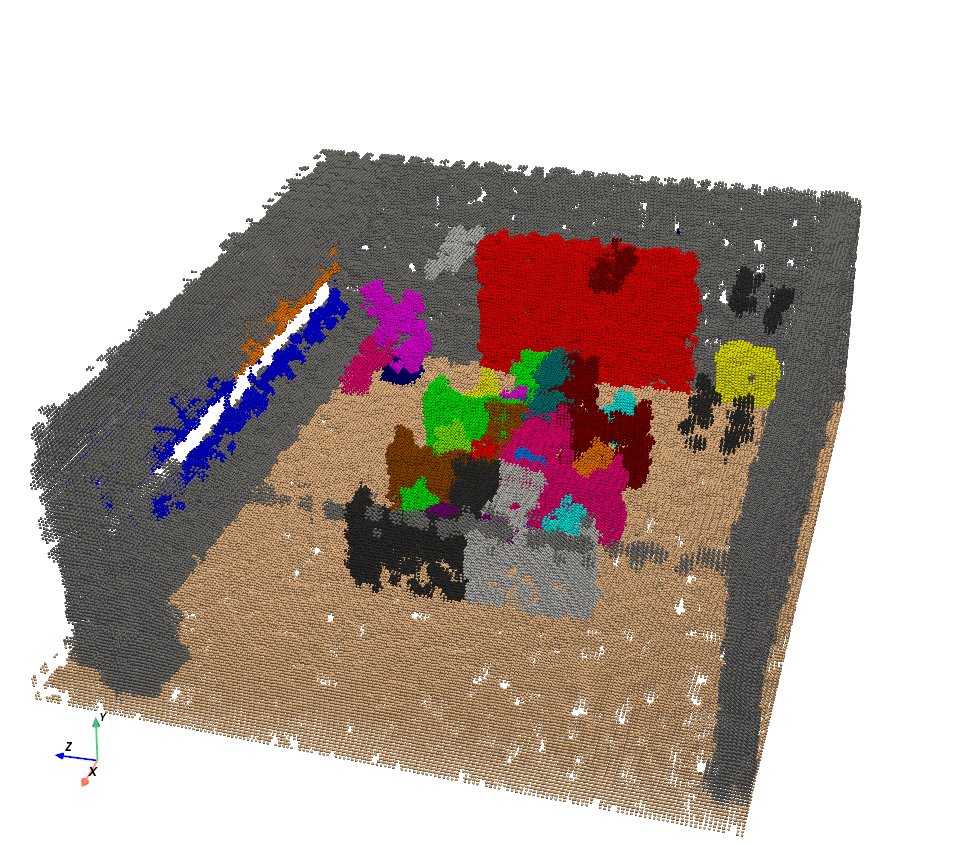}
        % \caption{}
        \label{fig:row1-rec-HMs}
    \end{subfigure}
    \begin{subfigure}{0.32\linewidth}
        \includegraphics[width=\linewidth]{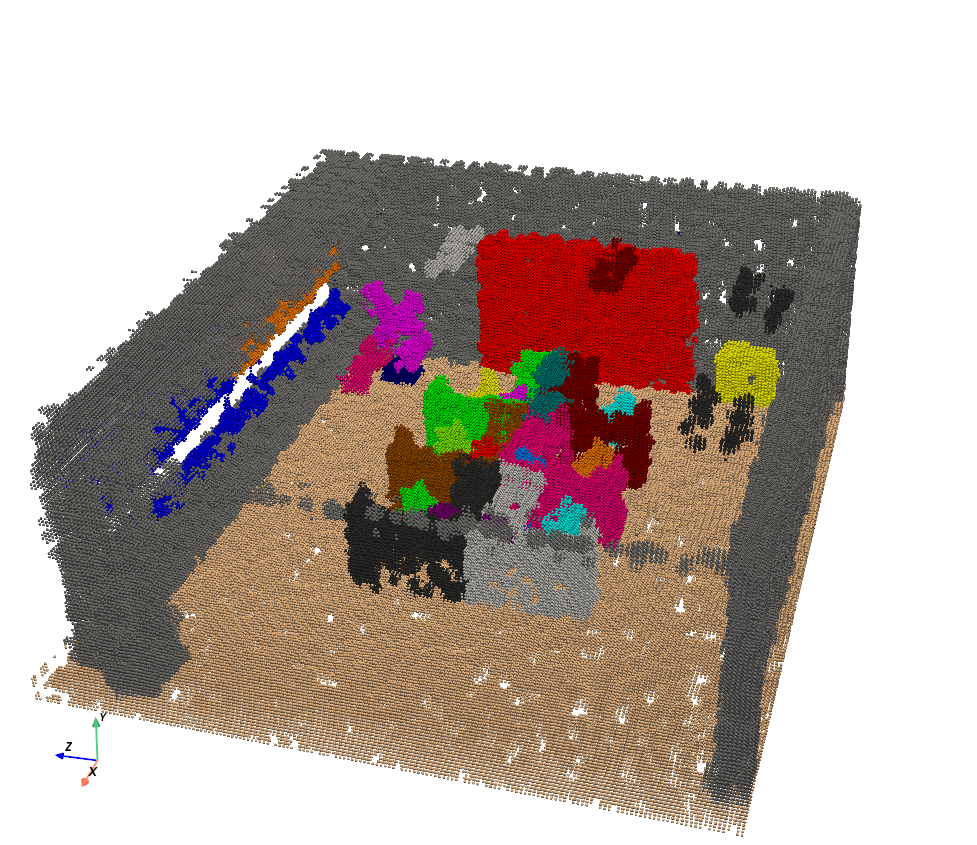}
        % \caption{}
        \label{fig:row1-rec-our}
    \end{subfigure}
    
    % \medskip % Vertical space between rows

    % --- Row 2 ---
    \begin{subfigure}{0.32\linewidth}
        \includegraphics[width=\linewidth]{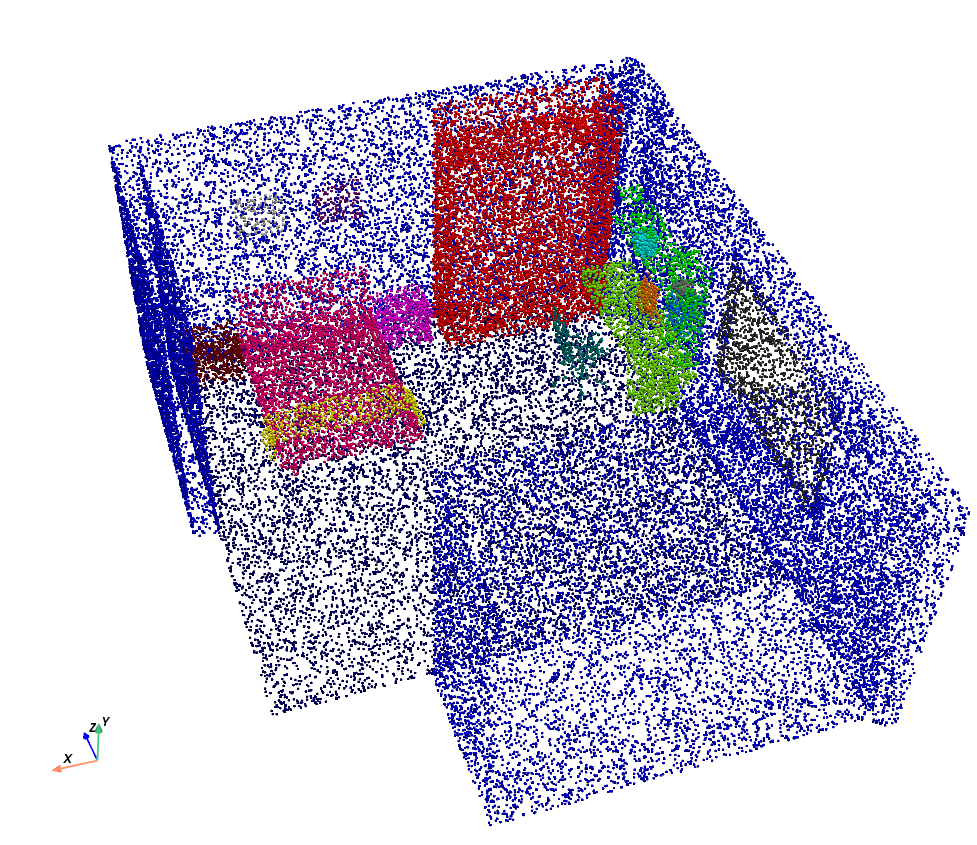}
        % \caption{}
        \label{fig:row2-gt}
    \end{subfigure}
    \begin{subfigure}{0.32\linewidth}
        \includegraphics[width=\linewidth]{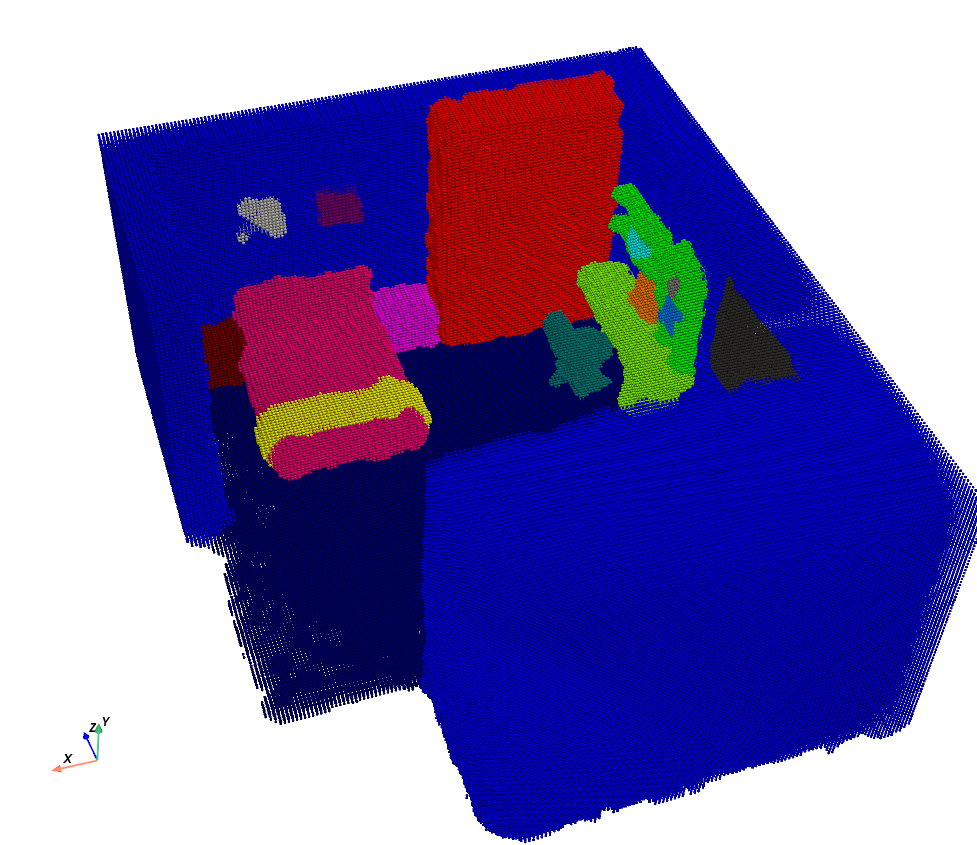}
        % \caption{}
        \label{fig:row2-rec-HMs}
    \end{subfigure}
    \begin{subfigure}{0.32\linewidth}
        \includegraphics[width=\linewidth]{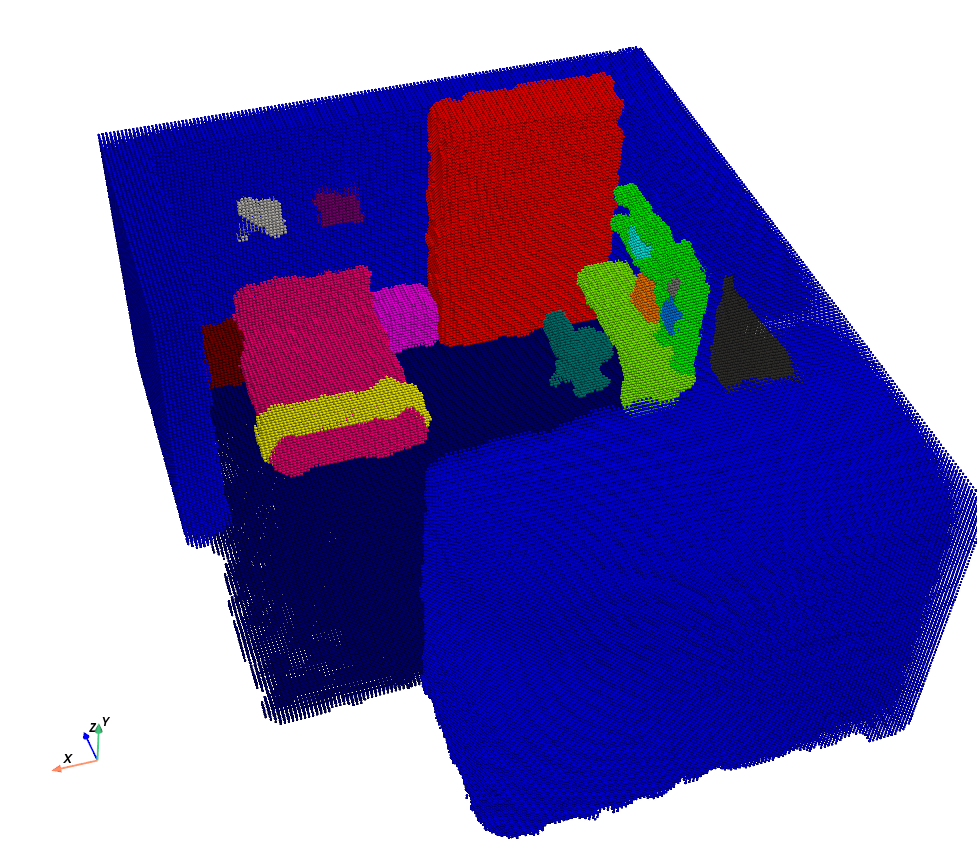}
        % \caption{}
        \label{fig:row2-rec-our}
    \end{subfigure}

    \caption{Reconstruction results on SceneNet dataset.}
    \label{fig:grid_comparison}
\end{figure}

\begin{figure}[t]
    \centering
    % --- Column Headers ---
    \begin{subfigure}{0.32\linewidth}
        \centering
        \textbf{Ground Truth}
    \end{subfigure}
    \begin{subfigure}{0.32\linewidth}
        \centering
        \textbf{HMs Results}
    \end{subfigure}
    \begin{subfigure}{0.32\linewidth}
        \centering
        \textbf{Our Results}
    \end{subfigure}
    
    % --- Row 1 ---
    \begin{subfigure}{0.32\linewidth}
        \includegraphics[width=\linewidth]{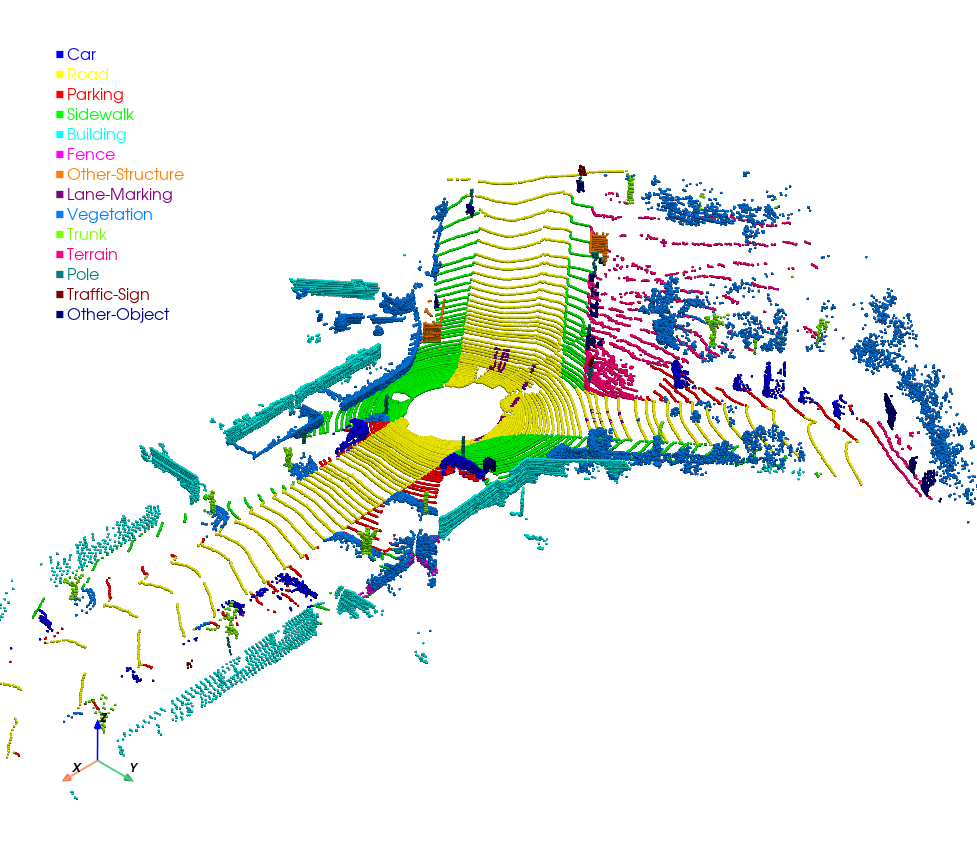}
        % \caption{}
        \label{fig:row1-gt}
    \end{subfigure}
    \begin{subfigure}{0.32\linewidth}
        \includegraphics[width=\linewidth]{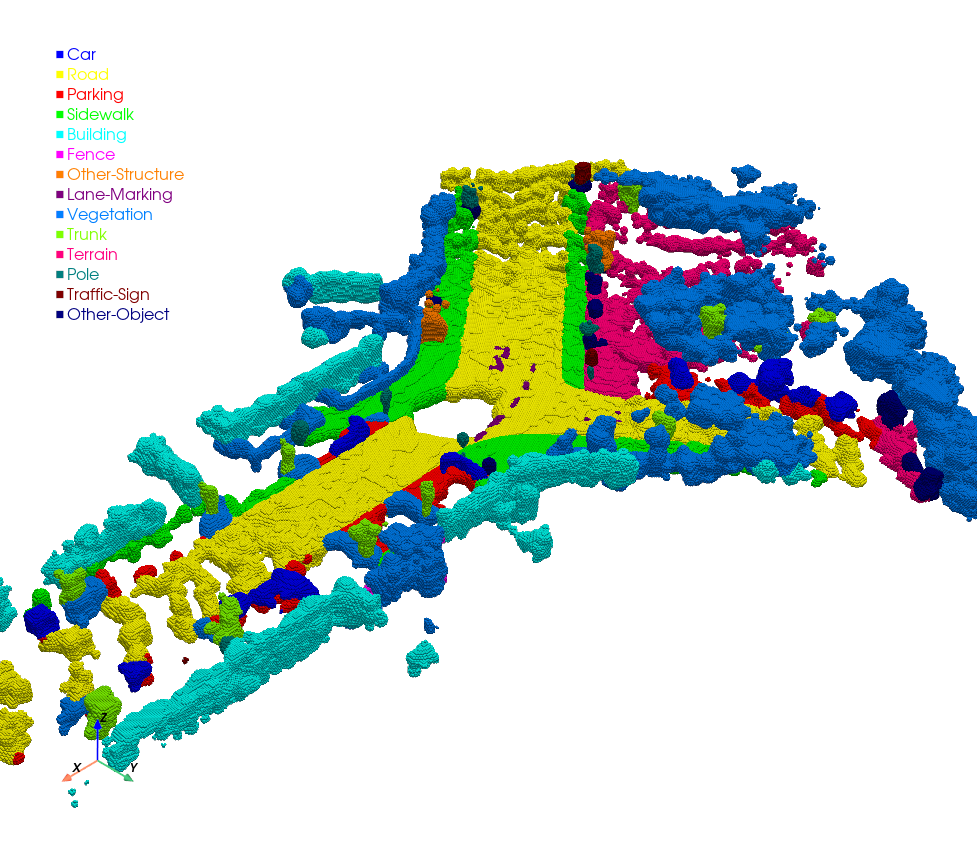}
        % \caption{}
        \label{fig:row1-rec-HMs}
    \end{subfigure}
    \begin{subfigure}{0.32\linewidth}
        \includegraphics[width=\linewidth]{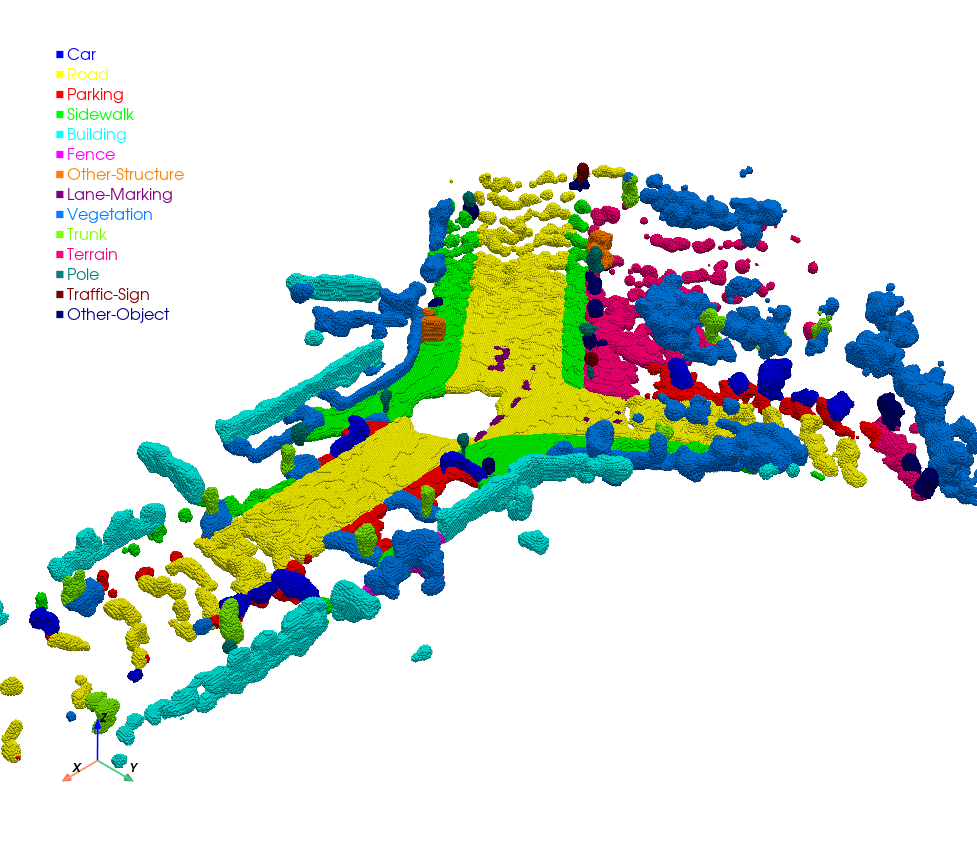}
        % \caption{}
        \label{fig:row1-rec-our}
    \end{subfigure}
    
    % \medskip % Vertical space between rows

    % --- Row 2 ---
    \begin{subfigure}{0.32\linewidth}
        \includegraphics[width=\linewidth]{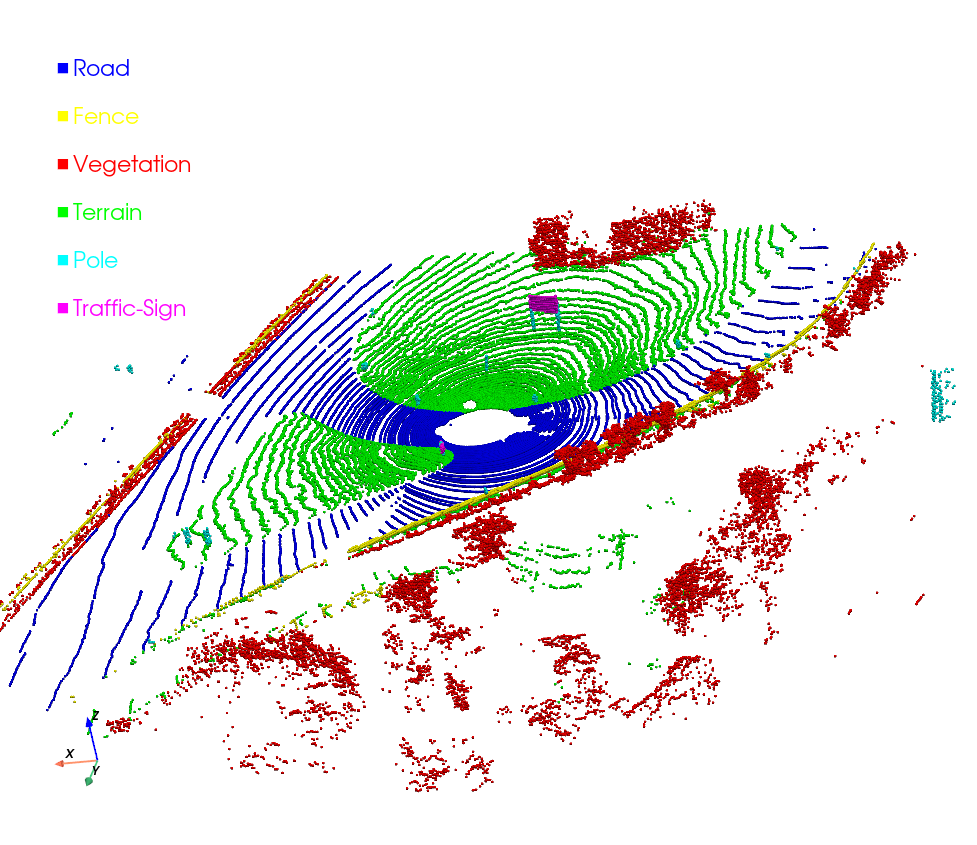}
        % \caption{}
        \label{fig:row2-gt}
    \end{subfigure}
    \hfill
    \begin{subfigure}{0.32\linewidth}
        \includegraphics[width=\linewidth]{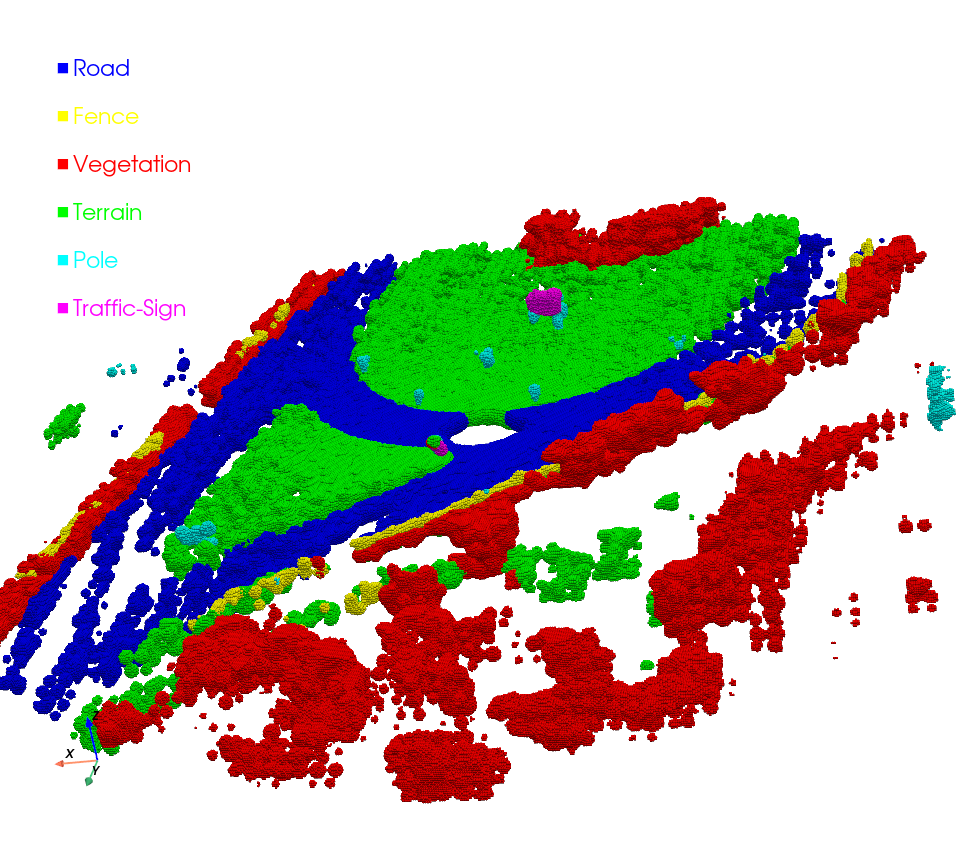}
        % \caption{}
        \label{fig:row2-rec-HMs}
    \end{subfigure}
    \hfill
    \begin{subfigure}{0.32\linewidth}
        \includegraphics[width=\linewidth]{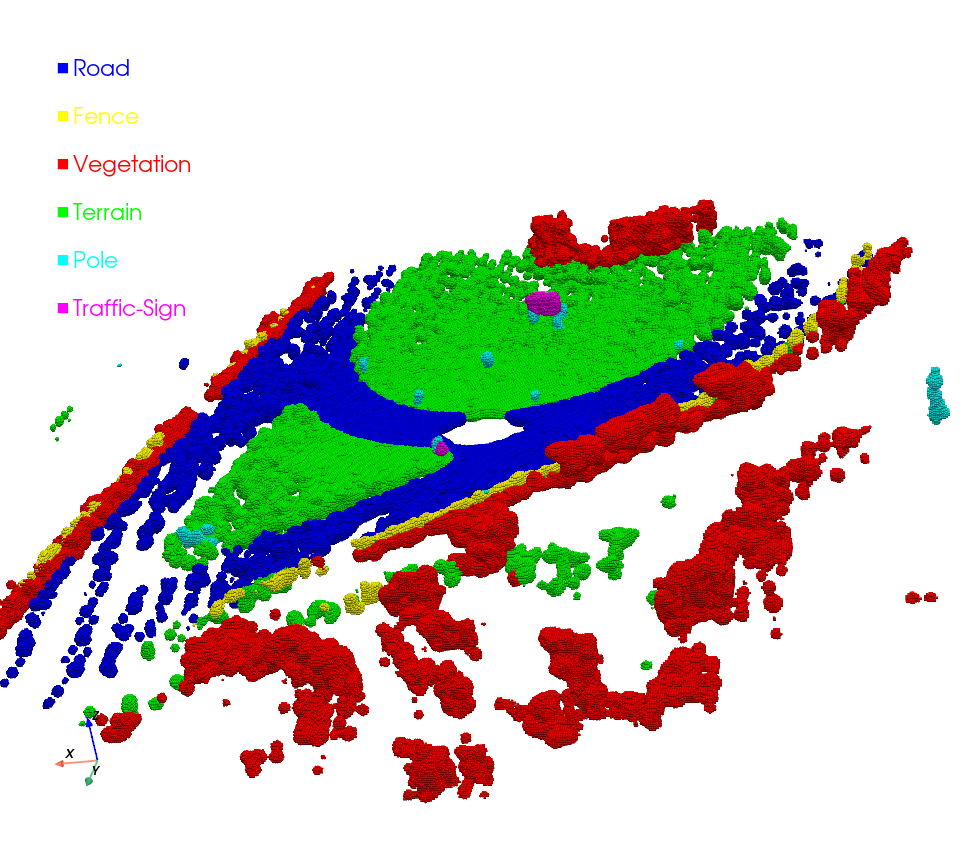}
        % \caption{}
        \label{fig:row2-rec-our}
    \end{subfigure}

    \caption{Reconstruction results on SemanticKITTI dataset.}
    \label{fig:grid_comparison2}
\end{figure}

Following the previous experiments, we further compare our method against 3D Hilbert Maps (HMs) \cite{3D_HilbertMaps} on two benchmarks, SemanticKITTI \cite{SemanticKITTI} and SceneNet \cite{SceneNet}, to evaluate performance in both large-scale outdoor and indoor environments. Experiments on SemanticKITTI are conducted using the initial scan of 10 distinct sequences, while SceneNet evaluation spans 60 independent indoor scenes (rooms). We report mIoU, training time, and inference time to assess not only semantic accuracy but also computational efficiency. In particular, this experiment aims to examine whether extending the original HM formulation with an additional uncertain output node, introduced to enable uncertainty-aware semantic mapping, can preserve or improve semantic accuracy while maintaining practical efficiency. Qualitative comparisons of semantic mapping results with 3D HMs are illustrated in Figures~\ref{fig:grid_comparison} and~\ref{fig:grid_comparison2}.

As shown in Tables \ref{tab:semantickitti} and \ref{tab:scenenet}, our method consistently achieves higher mIoU than 3D HMs on both datasets. This indicates that incorporating the "uncertain" class does not degrade semantic performance; instead, it leads to more accurate semantic predictions. A plausible explanation is that augmenting the training data with the "uncertain" class encourages the classifier to better separate in-distribution semantic classes from ambiguous or out-of-distribution regions. This results in clearer decision boundaries between positive classes and non-semantic or noisy observations, which is particularly beneficial in sparse or noisy 3D sensing scenarios.

In terms of time efficiency, our method incurs a small training-time overhead, while inference time remains nearly identical to that of 3D HMs. This overhead is expected, as both approaches employ a softmax classifier, but our model includes one additional output node to explicitly estimate uncertainty and needs to train on a larger augmented dataset including noise, which slightly increase the number of parameters and optimization cost. Nevertheless, the resulting training and inference times remain highly efficient and well within practical limits. In return for this marginal cost, our method enables uncertainty-aware semantic mapping, which is valuable for downstream tasks such as exploration, planning, and decision-making, while preserving and even improving semantic accuracy.

% -----------------------------------------------
% SemanticKITTI
% -----------------------------------------------

% =====================================================
\subsection{Compare to V-PRISM in representing tabletop scenes}
\begin{table}[t]
\centering
\caption{The performance of our proposed method compared to V-PRISM \cite{DBLP:conf/iros/WrightZJH24} on YCB dataset.}\label{tab:2}
\resizebox{\columnwidth}{!}{%
\begin{tabular}{|c|c|c|}
\hline
\textbf{}                            & \textbf{V-PRISM \cite{DBLP:conf/iros/WrightZJH24}} & \textbf{Our}                   \\ \hline
IoU $\uparrow$                       & 0.4994 ($\pm$ 0.0878)                              & 0.4854 ($\pm$ 0.0805)          \\
Chamfer (m) $\downarrow$             & 0.0126 ($\pm$ 0.026)                               & 0.0124 ($\pm$ 0.031)           \\
Training time (s) $\downarrow$       & 11.2826 ($\pm$ 5.6768)                             & \textbf{9.3533 ($\pm$ 2.1653)} \\
%Testing time (s) $\downarrow$       & 31.6787 ($\pm$ 10.9862)                            & \textbf{9.3518 ($\pm$ 1.8260)} \\
Reconstructing time (s) $\downarrow$ & 229.5563 ($\pm$ 100.0576)                          & \textbf{3.7560 ($\pm$ 1.1183)} \\ \hline
\end{tabular}
}
\end{table}

\begin{figure}[t]
    \centering
    \includegraphics[width=1.0\linewidth]{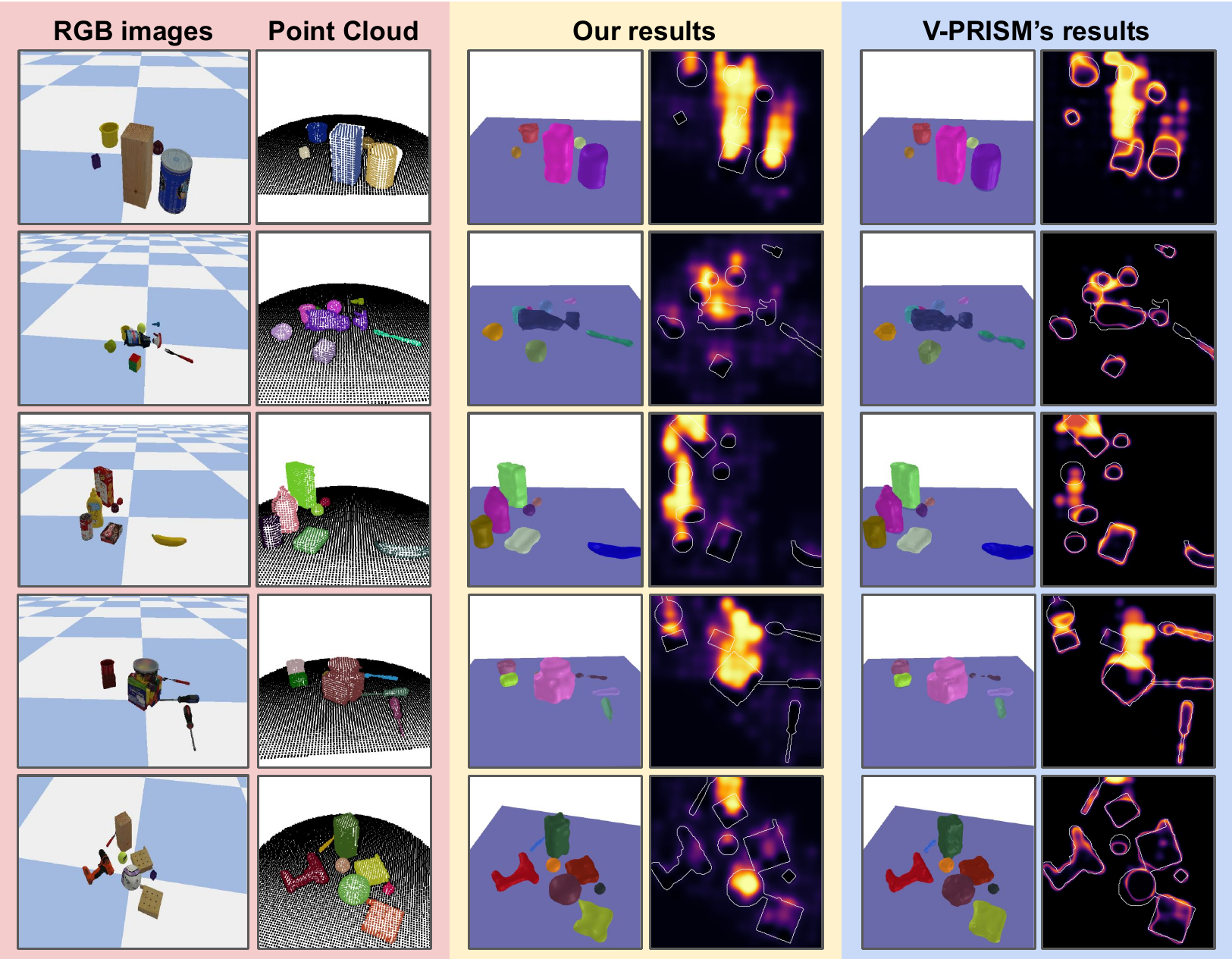}
    \caption{Representative scene reconstructions. Col. 1: RGB. Col. 2: segmented point cloud. Col. 3: our reconstructed mesh. Col. 4: our uncertainty (softmax auxiliary head; lighter = higher). Col. 5: V-PRISM mesh. Col. 6: V-PRISM uncertainty heatmap (lighter = higher). Following \cite{DBLP:conf/iros/WrightZJH24}, Cols. 4 and 6 show uncertainty on a 2D scene slice (bottom = nearer the camera). Object boundaries from the top-view projection are overlaid for spatial context.}
    \label{fig:reconstructed_scenes}
\end{figure}

\begin{figure}[t]
    \centering
    \includegraphics[width=1.0\linewidth]{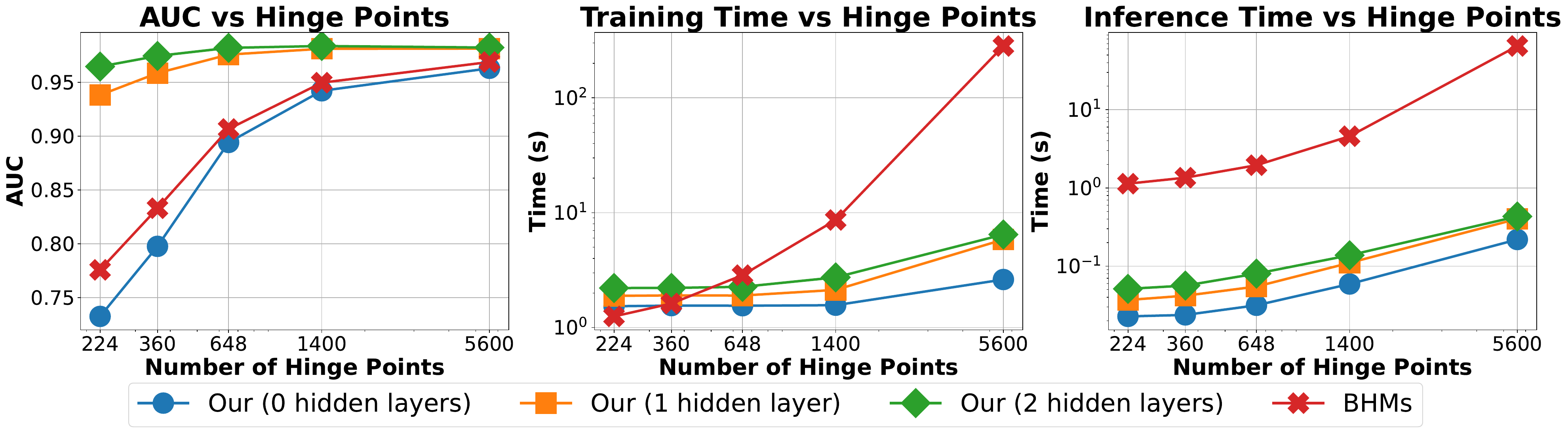}
    \caption{Effect of hinge point count on performance (Intel dataset). Results are shown for \textbf{our} methods with varying numbers of hidden layers and for \textbf{BHMs}.}
    \label{fig:hinge_points_effect}
\end{figure}

% \begin{figure}[t]
%     \centering
%     % --- Subfigure (a) AUC ---
%     \begin{subfigure}{0.32\linewidth}
%         \centering
%         \includegraphics[width=\linewidth]{figures/auc_vs_hingepoints_plot.pdf}
%         \caption{AUC}
%         \label{fig:hinge_auc}
%     \end{subfigure}
%     % --- Subfigure (b) Training time ---
%     \begin{subfigure}{0.32\linewidth}
%         \centering
%         \includegraphics[width=\linewidth]{figures/training_time_vs_hingepoints_plot.pdf}
%         \caption{Training time}
%         \label{fig:hinge_train}
%     \end{subfigure}
%     % --- Subfigure (c) Inference time ---
%     \begin{subfigure}{0.32\linewidth}
%         \centering
%         \includegraphics[width=\linewidth]{figures/inference_time_vs_hingepoints_plot.pdf}
%         \caption{Inference time}
%         \label{fig:hinge_infer}
%     \end{subfigure}

%     \caption{Effect of hinge point count on performance (Intel dataset). 
%     (\protect\subref{fig:hinge_auc}) Classification accuracy in terms of AUC, 
%     (\protect\subref{fig:hinge_train}) Training time, 
%     (\protect\subref{fig:hinge_infer}) Inference time. Results are shown for \textbf{our} methods with different hidden layers and for \textbf{BHMs}.}
%     \label{fig:hinge_points_effect}
% \end{figure}

In the next experiment, we investigate the ability of proposed method in handling the multiclass mapping problem in tabletop scenes scenario, and compare to the recent study V-PRISM \cite{DBLP:conf/iros/WrightZJH24}.
\noindent\textbf{Settings:} Following \cite{DBLP:conf/iros/WrightZJH24}, 100 scenes containing up to 10 objects each are generated from the YCB dataset \cite{YCB} to form the training and testing data. We also adopt the hinge-point generation mechanism of V-PRISM \cite{DBLP:conf/iros/WrightZJH24}, which has been shown to be effective for tabletop mapping.

When constructing the training set $\mathbf{D}$ for our method, we observed that sampling excessive noise in tabletop scenes degrades performance. Unlike occupancy mapping, the observed data are distributed across more classes; using the same amount of noise as in the previous setting would create a strong imbalance between sampled noise and samples in each class, hindering learning of the true data distribution. Therefore, we set the noise ratio to $2.5\%$ of the size of the actual training data. Additionally, due to the smaller noise ratio, noise samples are preferentially drawn near objects to increase the likelihood of capturing uncertain regions, such as occluded areas. Similar to \cite{DBLP:conf/iros/WrightZJH24}, the metrics used to assess performance are the intersection over union (IoU) and the Chamfer distance. We used the same process to calculate the Chamfer distance as in V-PRISM. In addition, we also report the average training and mesh reconstructing times of each method to compare their processing speed.

\noindent\textbf{Results Analysis:} The performance and runtime of our method and V-PRISM \cite{DBLP:conf/iros/WrightZJH24} are reported in Table \ref{tab:2}, while the third and fifth columns of Figure \ref{fig:reconstructed_scenes} illustrate representative examples of the reconstructed scenes. The results from Table \ref{tab:2} together with examples showed in Figure \ref{fig:reconstructed_scenes} state that our method achieves comparable accuracy to V-PRISM in representing the tabletop scene, with V-PRISM attaining a slightly higher average IoU score, whereas our method yields a marginally better Chamfer distance. However, in terms of time efficiency, our method surpasses V-PRISM, especially in scene reconstruction. This once again claims the substantial runtime advantage of our approach compared to a Bayesian-based method. With regard to uncertainty quantification, we show the result of the additional node, which corresponds to the "uncertain" class, in the fourth column of Figure \ref{fig:reconstructed_scenes}. The uncertain areas with respect to the occluded or partially observed sections of the scenes are highlighted correctly.

% =============================================
\subsection{Impact of hinge points count and Scaling ability of proposed method}

%  AUC score increase significantly with only a marginal rise of training and inference time. By contrast, BHM despite also have better accuracy, its training and inference time...
% stacking more hidden layer boost the performance with the cost of slightly increase runtime => show the potential of further improvement
In this part, we are going to empirically study the impact of hinge point count $H$, as well as the scalability of our method and Bayesian-based method when $H$ increases. We run occupancy mapping experiments on Intel dataset with different numbers of hinge points and recorded accuracy indicated by the AUC score, as well as the run time of our method and BHMs. The results are illustrated in Figure \ref{fig:hinge_points_effect}.

The results suggest that if we do not use enough hinge points to calculate features, or, in other words, if the number of features is insufficient, both methods cannot learn the environment landscape effectively. An increase in the number of hinge points leads to higher AUC scores. However, with BHMs, this comes at the cost of a dramatic rise in runtime, whereas, for our method, the runtime grows only modestly with the hinge point count. This emphasizes that our method scales much better than the Bayesian method, and is more practical to deploy in real world, especially for representing vast environments.
% stacking more hidden layer boost the performance with the cost of slightly increase runtime => show the potential of further improvement

As a softmax classifier can be considered as a lite version of a neural network with no hidden layers, we also carried out experiments which add 1 and 2 hidden layers to our model to investigate whether extending to an advanced architecture can substantially improve our method or not. The results reported in Figure \ref{fig:hinge_points_effect} indicate that adding more hidden layers can help the model perform exceptionally well even with only a small amount of initial features. This can be explained by the hidden layers provide the ability to learn hidden features in the middle of the training process, hence, reducing the need to start with a sufficient number of features at the input layer. It is worth noting that adding one or two hidden layers incurs only a slight increase in training and inference time. This observation supposes that our method can be further enhance for more accurate results, while still keeping the fast runtime and the uncertain estimation capability.
\section{Conclusion and Future Work}\label{sec:7}
This study proposes a straightforward yet effective approach for uncertainty estimation while maintaining accurate environment representation. Our method is inspired by noise contrastive estimation, utilizing noise samples as an additional "uncertain" class and learning it through a lightweight softmax classifier with an extra output node. Empirical results show that the proposed framework can produce accurate environment maps while simultaneously identifying uncertain regions, all with exceptionally low computational overhead. The uncertainty estimates produced by our model not only contribute to more reliable robot perception, but also enable active perception strategies, where robots prioritize exploration of poorly observed regions to improve mapping efficiency. In the future, we plan to investigate adaptive or scene-aware strategies for generating contrastive noise samples, as our experiments indicate that the choice of noise generation can significantly affect both the learned map and the resulting uncertainty estimation.

% \balance
\bibliographystyle{IEEEtran}
\bibliography{ref}
% \printbibliography
\end{document}